\documentclass[lettersize,journal]{IEEEtran}
\usepackage{amsmath,amsfonts}
\usepackage{algorithmic}
\usepackage{algorithm}
\usepackage{array}
\usepackage[caption=false,font=normalsize,labelfont=sf,textfont=sf]{subfig}
\usepackage{textcomp}
\usepackage{stfloats}
\usepackage{url}
\usepackage{verbatim}
\usepackage{graphicx}
\usepackage{booktabs}
\usepackage{cite}
\usepackage{multirow}
\usepackage[table]{xcolor}
\definecolor{DetBg}{RGB}{255,244,204}
\definecolor{ProbBg}{RGB}{226,240,218}
\newcommand{\detc}[1]{\cellcolor{DetBg}{#1}}
\newcommand{\probc}[1]{\cellcolor{ProbBg}{#1}}

\begin{document}

\title{FreCast: Refining Radar Echo Intensity via Phase-Preserving Amplitude Residual Diffusion for Precipitation Nowcasting}

% \author{Anonymous Authors}
\author{Heping Fang, Zihuai Yin, Kaicheng Mao, Peiguang Zhang and Peng Yang,~\IEEEmembership{Senior Member,~IEEE}
        % <-this % stops a space

\thanks{This work was supported in part by the National Natural Science Foundation of China under Grant 62272210, and Grant 62331014.(Corresponding author: Peng Yang).}
\thanks{Heping Fang is with Department of Statistics and Data Science, Southern University of Science and Technology, Shenzhen 518055, China (e-mail: 12331106@mail.sustech.edu.cn).}
\thanks{Zihuai Yin is with Department of Statistics and Data Science, Southern University of Science and Technology, Shenzhen 518055, China (e-mail: 12312003@mail.sustech.edu.cn).}
\thanks{Kaicheng Mao is with Department of Statistics and Data Science, Southern University of Science and Technology, Shenzhen 518055, China (e-mail: 12311704@mail.sustech.edu.cn).}
\thanks{Peiguang Zhang is with Department of Computer Science and Engineering, Southern University of Science and Technology, Shenzhen 518055, China (e-mail: 12431266@mail.sustech.edu.cn).}
\thanks{Peng Yang is with Department of Statistics and Data Science, Southern University of Science and Technology, Shenzhen 518055, China, and also with Guangdong Provincial Key Laboratory of Brain-Inspired Intelligent Computation, Department of Computer Science and Engineering and the Department of Statistics and Data Science, Southern University of Science and Technology, Shenzhen 518055, China (e-mail: yangp@sustech.edu.cn).}
}



\maketitle

\begin{abstract}

Precipitation nowcasting predicts the spatiotemporal evolution of future radar echoes from historical radar echo sequences, thereby estimating the occurrence, development, and movement of precipitation over the near term. In recent years, deep learning has become an important approach to precipitation nowcasting. Although state-of-the-art models can generally capture the overall spatial distribution of future precipitation, their predictions still exhibit substantial biases in radar echo intensity at individual locations. This observation motivates a more targeted strategy for reducing forecast errors. Instead of regenerating an entire radar echo sequence without spatial constraints, the predicted precipitation structure can be used to guide the refinement of echo intensities at individual locations. This structure-guided refinement directly targets echo intensity biases. Accordingly, we propose FreCast, a two-stage framework for radar echo prediction. The first stage generates an initial forecast of future radar echoes. The second stage uses the spatial structure of the initial forecast as a constraint to further correct intensity biases at individual locations in the first-stage prediction. Experiments on three datasets demonstrate that FreCast achieves consistent improvements across forecast skill metrics. Qualitative results further show that FreCast better preserves rainband continuity and intense precipitation structures at longer lead times.
\end{abstract}


\section{Introduction}
\IEEEPARstart{P}{recipitation} nowcasting seeks to predict the initiation, evolution and movement of precipitation systems over short lead time, typically 0--6h~\cite{pysteps,pan2024short}. Extreme precipitation often triggers floods, landslides and urban inundation, threatening human life and socioeconomic stability~\cite{nowcastnet,sun2014use,busker2025value,wang2024constructing}. Timely and reliable nowcasts are therefore needed for early warning and emergency response. 

Weather radar provides high-frequency, kilometer-scale observations of hydrometeor distributions across broad cloud and precipitation systems~\cite{han2021convective,dgmr}. Radar echo sequences can therefore reflect the evolution of precipitation. Unlike general video prediction, radar echo prediction does not primarily pursue perceptual realism. Instead, it places stricter requirements on the quantitative accuracy of echo intensity at each spatial location~\cite{li2025precipitation}. 
% This requires predicting the extent to which echo intensity increases or decreases at each location. These changes manifest as the intensification and decay of strong echo cores, echo initiation and dissipation, and the expansion and contraction of echo regions.

Traditional numerical weather prediction is constrained by update cycles, model resolution, and physical parameterizations, resulting in limited performance~\cite{sun2014use}. Radar echo extrapolation methods~\cite{li2004applications,wong2009towards,chen2023tempee} can predict echo displacement using estimated motion fields, but they struggle to capture intensity changes that cannot be explained by horizontal motion~\cite{jing2022remnet}. In recent years, deep learning methods have improved radar echo prediction through end-to-end spatiotemporal modeling and have become an important approach to precipitation nowcasting.

% Most existing deep learning methods formulate precipitation nowcasting as a full-frame image sequence prediction task. Deterministic models based on convolutional recurrent networks, Transformers and their variants can learn dominant motion patterns of radar echoes and achieve high pixel-level similarity at short lead times~\cite{convlstm,kim2025hybrid,earthformer,dong2022motion}. However, these models are usually trained by directly minimising the distance between predicted and ground-truth frames, which can produce over-smoothed predictions and underestimate peak intensities in strong echo regions. By contrast, generative methods based on generative adversarial networks or diffusion models can recover richer local textures and details~\cite{dgmr,prediff,luo2022experimental,duocast}. Yet the high degrees of freedom in pixel-domain generation may lead to spatial displacement, structural deformation, boundary fragmentation or spurious echoes. These two modeling paradigms therefore produce distinct error patterns. Rather than attributing such errors mainly to insufficient spatiotemporal modeling capacity, we re-examine existing nowcasting models by asking where their prediction errors arise.

Most existing deep learning approaches to precipitation nowcasting predict future radar echoes directly in the pixel domain and can be broadly divided into deterministic and generative paradigms. Deterministic forecasting models~\cite{convlstm,kim2025hybrid,earthformer,dong2022motion} map an observed radar sequence to a single future sequence and are commonly trained using pixelwise reconstruction losses. These models tend to produce overly smooth predictions and underestimate the peak intensities of strong echoes. In contrast, generative forecasting models~\cite{dgmr,prediff,luo2022experimental,duocast} learn the conditional distribution of pixel-level future echoes and can recover visually richer local details, but they do not necessarily predict radar echo intensity fields accurately, with the resulting errors manifesting as echo fragmentation, spatial displacement, and structural deformation. These two modeling paradigms therefore exhibit distinct error patterns. Rather than attributing these errors primarily to architectural limitations of different deep learning models, we re-examine existing nowcasting models by investigating where their prediction errors arise.

\begin{figure}[htbp]
\centerline{\includegraphics[width=\columnwidth]{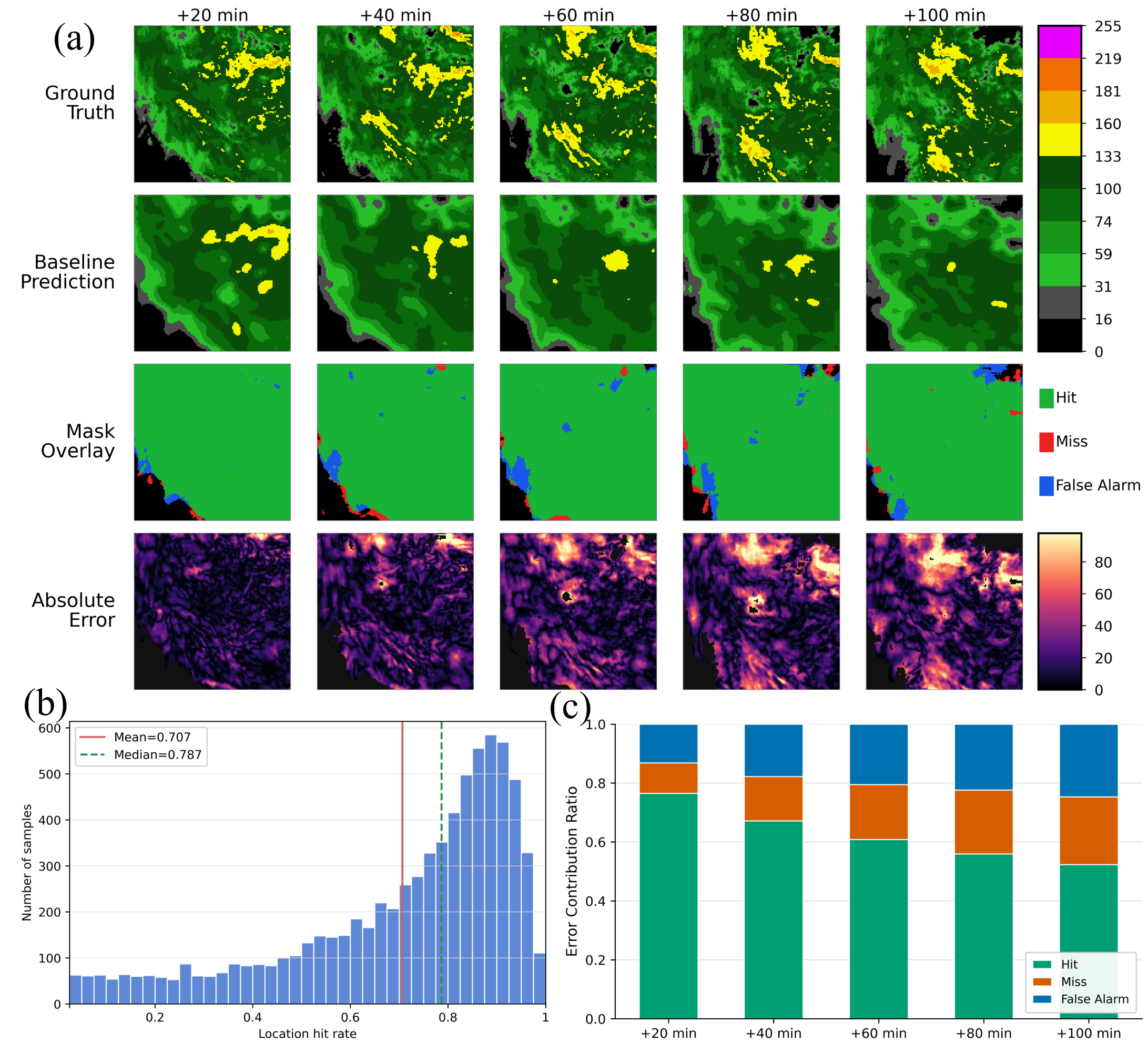}}
\caption{Analysis of error sources in radar echo nowcasting. At each spatial location, \textbf{\emph{hit}} denotes an echo in both the prediction and ground truth, \textbf{\emph{miss}} denotes no predicted echo where the ground truth contains one, and \textbf{\emph{false alarm}} denotes a predicted echo where the ground truth contains none.
\textbf{(a)}, A SEVIR test example comparing valid echo regions predicted by the AlphaPre baseline model with the corresponding ground-truth regions, together with categorical mask overlays and absolute error maps. Despite a hit rate of 0.9882, substantial echo intensity errors remain within the hit region.
\textbf{(b)}, Distribution of hit rates over the test set, with mean and median values of 0.707 and 0.787, respectively, indicating that the prediction errors are not primarily caused by large-scale spatial displacement.
\textbf{(c)}, Contributions of the hit, missed, and false alarm regions to the total absolute error at different forecast lead times. The hit region, shown in green, consistently accounts for the largest proportion of the total error.}
\label{fig:fig_insight}
\end{figure}

We evaluate AlphaPre~\cite{alphapre}, a recently proposed representative model for precipitation nowcasting, on the Storm Event Imagery (SEVIR) dataset~\cite{sevir} and analyze its predictions. Fig.~\ref{fig:fig_insight} uses AlphaPre as an illustrative example to visualize the sources of error in radar echo forecasts. The results show that AlphaPre can locate the main precipitation regions with reasonable accuracy, while a substantial proportion of the remaining error arises from inaccurate estimates of echo intensity within the hit regions. The qualitative forecasts of six representative baseline models shown later in Fig.~\ref{fig:p0_methods_prediction_visualization_compare} also exhibit similar intensity estimation errors. These preliminary observations inspire us to explore whether forecast performance can be further improved by correcting echo-intensity biases within predicted precipitation regions.

To specifically correct echo-intensity biases, we formally interpret the evolution of future radar echoes as the joint effect of advective transport and non-advective intensity change. Under this formulation, we adopt a two-stage predict-then-refine strategy. This strategy separately models the baseline forecast of the future echo field and the correction of its remaining intensity biases. 
Existing two-stage nowcasting methods typically first generate a deterministic baseline forecast in the pixel domain. They then learn a stochastic residual in the same domain to compensate for variations not fully captured by the baseline~\cite{diffcast,cascast}. From the intensity evolution perspective adopted in this study, this process also can be interpreted as a form of intensity-field decomposition. The resulting pixel-domain residual is typically defined as the pixelwise difference between the baseline forecast and the ground truth. However, it contains both spatial-structure errors and echo-intensity errors. Consequently, a pixel-domain refinement model may alter the precipitation structure predicted in the first stage while recovering local intensity details. To impose a more explicit spatial constraint on refinement, we formulate the second-stage intensity correction in the frequency domain. Previous research has shown
that phase variations in the frequency domain are strongly associated with the
changes in the locations and morphologies of precipitation
regions, whereas amplitude variations are strongly associated with
changes in precipitation intensity~\cite{oppenheim2005importance,alphapre,yan2024fourier}. Although amplitude and phase do not perfectly separate structure from intensity, they provide useful proxy representations for constraining the refinement process.

Based on this insight, we propose \textbf{FreCast}, a two-stage framework for radar echo prediction. In the first stage, FreCast predicts the future amplitude and phase spectra and reconstructs an initial forecast. In the second stage, FreCast retains the predicted phase as a spatial-structure anchor and models only the remaining amplitude-spectrum residual. The supervision target is defined as the difference between the normalized ground-truth amplitude spectrum and the baseline amplitude spectrum. This design focuses the generative correction on intensity-related spectral errors while limiting structural changes introduced during refinement.

The main contributions of this study are summarized as follows:
\begin{enumerate}
    \item We conduct an illustrative error-source analysis of an AlphaPre forecast for a case in the SEVIR dataset. This preliminary observation motivates us to investigate whether targeted correction of echo intensity biases can further improve nowcasting performance when the overall spatial structure has been predicted with reasonable accuracy.

    \item We introduce a structurally constrained frequency-domain refinement strategy for two-stage precipitation nowcasting. The strategy retains the phase predicted in the first stage as a spatial-structure anchor and models only an explicitly supervised amplitude-spectrum residual.

    \item We develop FreCast to implement this two-stage modeling strategy. Experiments on multiple radar echo datasets demonstrate that FreCast reduces echo intensity biases and improves nowcasting performance.
\end{enumerate}

The remainder of this paper is organized as follows. Section~\ref{sec:related_work} and Section~\ref{sec:motivation} review the related work and present the motivation, respectively. Section~\ref{sec:methodology} describes FreCast, Section~\ref{sec:experiments_and_analyses} reports the experiments, and Section~\ref{sec:conclusion} concludes the paper and discusses future research directions.

\section{Related Works}
\label{sec:related_work}

Existing deep learning-based methods for precipitation nowcasting can be reviewed along two complementary dimensions. These dimensions concern the forecasting paradigm and the modeling domain. With respect to the forecasting paradigm, existing methods can be broadly divided into deterministic and probabilistic generative approaches. With respect to the modeling domain, recent studies have extended prediction from the pixel domain to the frequency domain. Following this organization, we first review deterministic forecasting methods. We then discuss probabilistic generative forecasting and residual refinement. Finally, we position FreCast with respect to the most closely related approaches.

\subsection{Deterministic Spatiotemporal Predictive Learning for Precipitation Nowcasting}
Based on how temporal dependencies are modeled, deterministic precipitation nowcasting methods can be grouped into recurrent and non-recurrent approaches. ConvLSTM is a representative recurrent method. PredRNN, MIM, PredRNN++, and MAU further improve memory propagation, nonstationary dynamics modeling, and motion representation~\cite{convlstm,predrnn,mim,predrnn++,mau}. Non-recurrent methods employ convolutional encoder--translator--decoder architectures or attention mechanisms. These designs improve computational parallelism and enlarge spatiotemporal receptive fields. Representative methods include SimVP, TAU, EarthFormer, and Rainformer~\cite{simvp,tau,earthformer,rainformer}. Despite their architectural differences, these models are commonly trained with pixelwise reconstruction objectives. They capture dominant motion trends and large-scale echo organization. Thus, they also serve as important baselines for subsequent refinement methods. However, when future evolution is multimodal, regression objectives may favor the conditional mean. At longer lead times, this tendency may lead to degradation of the detail and underestimation of strong echoes.

\subsection{Probabilistic Spatiotemporal Predictive Learning and Residual Refinement for Precipitation Nowcasting}
Based on their modeling targets, probabilistic generative methods for precipitation nowcasting can be divided into direct full-field generation and deterministic forecast-guided refinement. Methods in the first group directly learn the conditional distribution of complete future radar echo sequences. DGMR~\cite{dgmr} employs conditional adversarial generation. It constrains the generated sequences using spatial and temporal discriminators. PreDiff~\cite{prediff} instead performs conditional diffusion in a compressed latent space. A knowledge alignment mechanism guides the generation process. Methods in the second group first produce a physics-constrained or deterministic intermediate prediction. This prediction conditions the generation of local variations that the deterministic forecast does not fully capture. NowcastNet~\cite{nowcastnet} constrains its generative network using the output of a differentiable physical evolution network. DiffCast~\cite{diffcast} models the residual in the pixel-domain between a deterministic prediction and the ground truth using a temporally conditioned diffusion process. CasCast~\cite{cascast} cascades deterministic mesoscale prediction with probabilistic modeling of small-scale variations in a latent space. Compared with direct full-field generation, these methods condition probabilistic generation on a structured intermediate prediction. However, they differ in their refinement targets and modeling domains.

\subsection{Frequency-Domain Modeling for Precipitation Nowcasting}
Frequency information has been incorporated into precipitation nowcasting through training objectives, predictive representations, and frequency-decomposed stochastic modeling. FACL~\cite{yan2024fourier} replaces conventional objectives of type \(L_2\) with Fourier amplitude and correlation losses. It introduces spectral constraints at the optimization level without changing the full-field prediction task. AlphaPre~\cite{alphapre} introduces frequency decomposition at the representation level. It separately predicts future amplitude and phase representations. AlphaMixer then fuses these representations to reconstruct the complete future echo sequence deterministically. DuoCast~\cite{duocast} uses a different decomposition strategy. It projects the complete future field into low frequency and high frequency target components. Separate stochastic diffusion processes model the two components. The high-frequency stage is conditioned on the low-frequency prediction and refines local variations in a latent space. The role of frequency information therefore depends on where it is introduced into the forecasting pipeline. It can provide spectral supervision or support decomposed predictive modeling.

\subsection{Positioning of FreCast}
FreCast follows a two-stage paradigm in which baseline prediction is followed by residual correction. Under this paradigm, the most closely related methods are AlphaPre, DiffCast, and DuoCast. AlphaPre deterministically learns and fuses future amplitude and phase representations to reconstruct the complete future echo sequence. DiffCast uses a deterministic prediction as its baseline. It applies diffusion to the pixel-domain residual between the ground truth and this baseline. DuoCast projects the complete echo field into low- and high-frequency components. Separate low- and high-frequency diffusion models represent the two components. The high-frequency model is conditioned on the low-frequency prediction. By contrast, FreCast restricts second-stage stochastic refinement to the amplitude-spectrum residual of a deterministic spectral baseline. During spectral recombination, it reuses the predicted baseline phase and does not sample a phase residual.

\section{Motivation}
\label{sec:motivation}

\subsection{From Forecast Error Analysis to Intensity-Field Evolution Modeling}

Precipitation nowcasting is commonly formulated as an image sequence prediction problem. Given historical radar echo observations \(X=\{X_t\}_{t=-T_{\mathrm{in}}+1}^{0}\), with \(X\in\mathbb{R}^{T_{\mathrm{in}}\times C\times H\times W}\), a model learns a mapping function \(\mathcal{F}_{\theta}\) to predict the future radar echo sequence \(Y=\{Y_t\}_{t=1}^{T_{\mathrm{out}}}\), with \(Y\in\mathbb{R}^{T_{\mathrm{out}}\times C\times H\times W}\). This formulation can be written as
\begin{equation}
Y
=
\mathcal{F}_{\theta}(X).
\label{eq:seq-mapping}
\end{equation}

According to the error-source analysis in Section~I, pronounced echo-intensity biases within precipitation hit regions constitute a major source of forecast error. Radar pixel values encode quantitatively meaningful echo intensities. This error pattern is therefore difficult to interpret clearly from the perspective of image sequence prediction alone. To distinguish the motion of spatial echo structures from local changes in echo intensity, we regard radar sequences as discrete observations of a spatiotemporal intensity field. We then analyze future echoes from the perspective of intensity-field evolution.

Each radar echo frame can be represented as an intensity field
\begin{equation}
\mathbf{I}_{t}
=
\left\{
I(p,t)
\mid
p\in\Omega
\right\},
\label{eq:intensity-field}
\end{equation}
where \(p\) denotes a two-dimensional spatial location and \(I(p,t)\) denotes the radar echo intensity at location \(p\) and time \(t\). Accordingly, we formulate the prediction of intensity-field evolution as
\begin{equation}
I(p,t+\ell)
=
\mathcal{F}_{\theta}
\left(
\left\{
I(q,\tau)
\ \middle|\
q\in\Omega,\,
\tau
\right\}
\right),
\label{eq:intensity-evolution}
\end{equation}
where \(\tau\) indexes the historical time steps within the input window, i.e., $\tau=t-T_{\rm in}+1,\ldots,t$. The variable \(\ell\) denotes the forecast lead time and satisfies \(\ell=1,\ldots,T_{\mathrm{out}}\). This formulation requires the model to estimate the amount by which echo intensity increases or decreases at each spatial location. Such changes manifest as the intensification and decay of strong echo cores, as well as echo initiation and dissipation.

\subsection{Decomposition of Intensity Fields}

From the intensity-field perspective, radar echo evolution can be conceptually viewed as the combined effect of advective transport and nonadvective processes. The corresponding continuous form can be expressed through a transport--source relation~\cite{nowcastnet}
\begin{equation}
\frac{\partial I}{\partial t}
+
\mathbf{u}\cdot\nabla I
=
S,
\label{eq:transport-source}
\end{equation}
where \(\mathbf{u}\) denotes the echo motion field. The term \(\mathbf{u}\cdot\nabla I\) describes the advective transport of existing echo structures. The term \(S\) represents local source--sink effects that cannot be explained by advection alone. Within a finite forecast horizon, the future intensity field can be conceptually approximated as
\begin{equation}
I(p,t+\ell)
\approx
I_{\mathrm{adv}}(p,t+\ell)
+
I_{\mathrm{chg}}(p,t+\ell),
\label{eq:adv-chg-decomposition}
\end{equation}
where \(I_{\mathrm{adv}}\) denotes the background echo field produced by advective transport and \(I_{\mathrm{chg}}\) denotes the signed local nonadvective intensity change. This decomposition has a clear physical interpretation, but its components are not directly available as supervised learning targets. Standard radar echo datasets provide only the observed intensity fields \(I(p,t+\ell)\). Separate labels for the true advective component \(I_{\mathrm{adv}}\) and the local change component \(I_{\mathrm{chg}}\) are unavailable. Existing models can therefore only approximate this decomposition indirectly.

More specifically, full-field forecasting methods include ConvLSTM~\cite{convlstm}, SimVP~\cite{simvp}, DGMR~\cite{dgmr}, and PreDiff~\cite{prediff}. These methods can be viewed as directly learning the future echo field jointly determined by \(I_{\mathrm{adv}}\) and \(I_{\mathrm{chg}}\) through a holistic black-box mapping \(\hat{Y}=\mathcal{F}_{\theta}(X)\). A second class of methods uses structured intermediate forecasts to guide subsequent probabilistic generation. NowcastNet~\cite{nowcastnet}, DiffCast~\cite{diffcast}, and CasCast~\cite{cascast} introduce different forms of intermediate forecast conditions and then model residual corrections. Their generic formulation can be written as \(\hat{Y}=q_{\phi}(X)+\epsilon_{\psi}\left(X,q_{\phi}(X),z\right).\) Here, \(q_{\phi}(X)\) produces a deterministic forecast. The variable \(z\sim p_z(z)\) denotes random noise or a latent variable used to characterize uncertainty in future evolution. The function \(\epsilon_{\psi}(\cdot)\) generates the residual correction term.

When forecast errors are analyzed in the pixel domain, the prediction error associated with a forecast \(\hat{I}(p,t+\ell)\) can be written as\(e(p,t+\ell)=\hat{I}(p,t+\ell)-I(p,t+\ell).\) This error can arise from different factors. A spatial displacement of the predicted echo structure can produce a large pixel error even when the predicted intensity is reasonable. Conversely, a substantial error can also arise when the echo location is approximately correct but its intensity is underestimated or overestimated. The pixel-domain error can therefore be conceptually written as
\begin{equation}
e(p,t+\ell)
\approx
e_{\mathrm{loc}}(p,t+\ell)
+
e_{\mathrm{int}}(p,t+\ell),
\label{eq:error-decomposition}
\end{equation}
where \(e_{\mathrm{loc}}(p,t+\ell)\) denotes the location-related error and \(e_{\mathrm{int}}(p,t+\ell)\) denotes the intensity-related error. These two components are difficult to separate in the original pixel domain because spatial location and echo intensity jointly determine each pixel value. Consequently, directly learning the full mapping \(\hat{Y}=\mathcal{F}_{\theta}(X)\) or directly modeling the pixel residual \(\Delta Y=Y-\hat{Y}\) can mix spatial-structure biases and intensity biases within a single correction target.

\subsection{Amplitude-Spectrum Residual Refinement Under a Baseline Phase Constraint}

To obtain a more explicitly constrained proxy representation, we transform the radar echo intensity field into the frequency domain. For a future radar echo frame \(\mathbf{I}_{t+\ell}\), its two-dimensional Fourier transform can be expressed as
\begin{equation}
\mathbf{Z}_{t+\ell}
=
\operatorname{FFT}
\left(
\mathbf{I}_{t+\ell}
\right)
=
\mathbf{A}_{t+\ell}
\odot
\exp
\left(
i\boldsymbol{\Phi}_{t+\ell}
\right),
\label{eq:fft-amp-phase}
\end{equation}
where \(\mathbf{A}_{t+\ell}=|\mathbf{Z}_{t+\ell}|\) and \(\boldsymbol{\Phi}_{t+\ell}=\arg(\mathbf{Z}_{t+\ell})\) denote the amplitude spectrum and phase spectrum, respectively. The symbol \(\odot\) denotes the Hadamard product. Amplitude and phase are not exact or independent representations of echo intensity and spatial structure. Nevertheless, compared with pixel-domain residual decomposition, they can be regarded as proxy representations with greater relative sensitivity to intensity-related and structure-related changes, respectively.

Motivated by this perspective, we formulate precipitation nowcasting as a frequency-domain amplitude-residual refinement task under a baseline phase constraint. This formulation uses the spatial-structure prior contained in the initial forecast while providing a more targeted direction for correcting local intensity errors.

\section{Methodology}
\label{sec:methodology}

\subsection{Overview}

\begin{figure*}[!t]
\centering
\includegraphics[width=\textwidth]{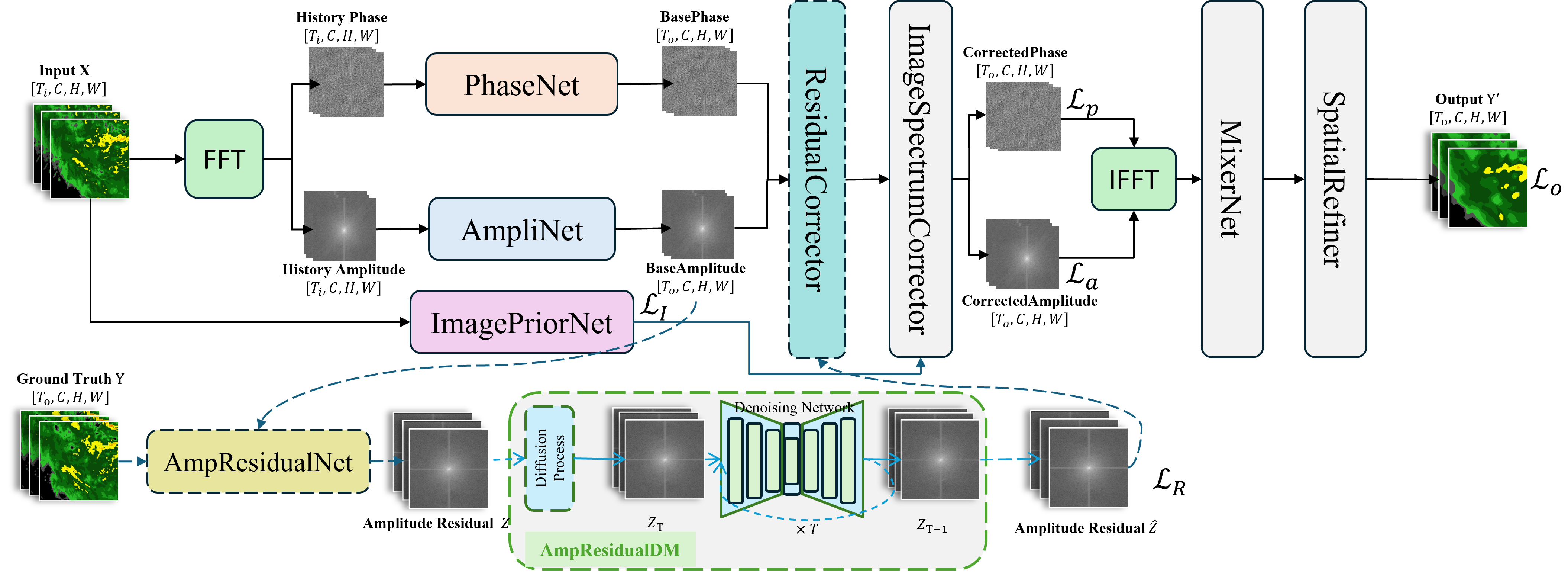}
\caption{Overall framework of FreCast. Given a sequence of historical radar echo frames, the backbone first predicts the future amplitude and phase spectra in the frequency domain and reconstructs a structured deterministic baseline forecast. The amplitude residual diffusion model then generates the remaining amplitude residual conditioned on the baseline amplitude and low-frequency phase structure. Finally, the sampled amplitude residual is added back to the baseline amplitude, while the phase is retained as a spatial structural anchor. The final precipitation nowcast is obtained through inverse transformation.}
\label{fig:frecast_framework}
\end{figure*}

As shown in Fig.~\ref{fig:frecast_framework}, FreCast consists of two stages. Section IV-B presents the deterministic spectral forecasting stage. AmpliNet and PhaseNet predict the amplitude and phase, respectively, while ImagePriorNet performs image-prior-guided spectral correction. MixerNet and SpatialRefiner subsequently perform spectrum-to-image reconstruction. Section IV-C introduces the conditional amplitude-residual diffusion network, AmpResidualDM, and phase-preserving reconstruction. Section IV-D describes the two-stage training procedure and the corresponding inference process.

\subsection{Deterministic Spectral Baseline}

The deterministic backbone provides the structured spectral reference on which amplitude residual correction is performed. For each historical radar frame \(\mathbf{I}_{\tau}\), where \(\tau=t-T_{\mathrm{in}}+1,\ldots,t\), we compute \(\mathbf{Z}_{\tau}=\mathrm{FFT}(\mathbf{I}_{\tau})\), with \(\mathbf{A}_{\tau}=|\mathbf{Z}_{\tau}|\) and \(\boldsymbol{\Phi}_{\tau}=\arg(\mathbf{Z}_{\tau})\). Here, \(\mathbf{A}_{\tau}\) describes the frequency-energy distribution, while \(\boldsymbol{\Phi}_{\tau}\) carries spatial alignment information. The backbone predicts these two components with separate but structurally similar branches.

\subsubsection{Amplitude and Phase Prediction}

\begin{figure}[t]
\centering
\includegraphics[width=\columnwidth]{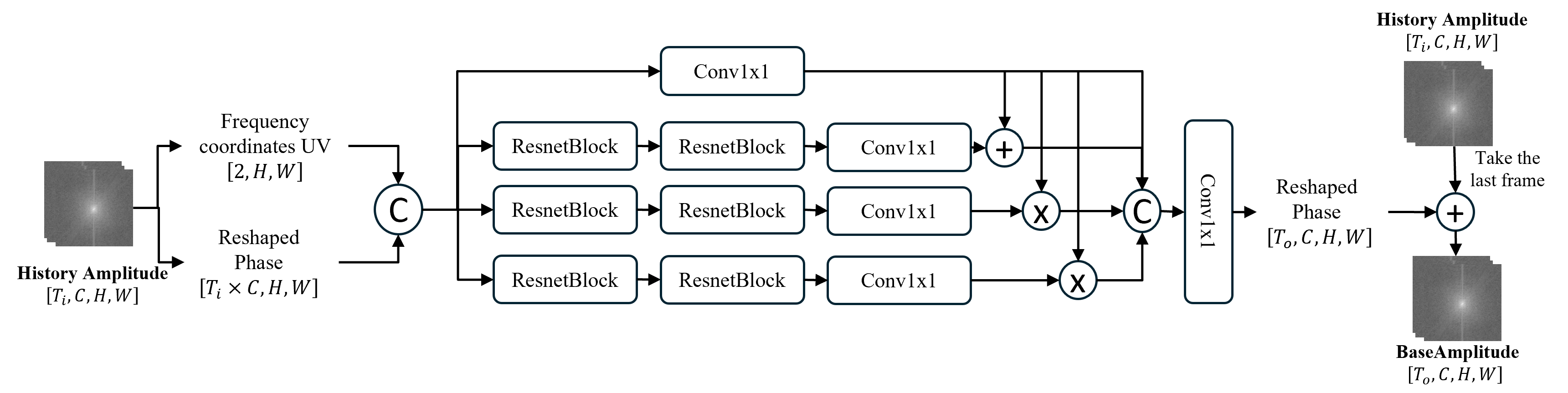}
\caption{Amplitude spectrum prediction branch. AmpliNet learns the evolution of future amplitude spectra and provides the intensity basis for subsequent reconstruction. PhaseNet adopts the same prediction structure but differs in its input and output representations.}
\label{fig:amplinet}
\end{figure}

As shown in Fig.~\ref{fig:amplinet}, AmpliNet predicts the future amplitude spectra from historical amplitude information and frequency coordinates. Its input is \(\mathbf{H}_{A}=\operatorname{Concat}(\mathcal{N}_{A}(\mathbf{A}_{\tau}),\mathbf{U})\), where \(\mathcal{N}_{A}(\cdot)\) denotes the amplitude normalization operator and \(\mathbf{U}\) denotes the two-dimensional frequency-coordinate map.

PhaseNet uses the same multi-branch prediction principle as AmpliNet, but its input is adapted to the periodic nature of phase. It takes sine and cosine phase representations, amplitude information, and frequency coordinates as inputs
\begin{equation}
\begin{aligned}
\mathbf{H}_{\Phi}
=
\operatorname{Concat}
\bigl(
&\sin(\boldsymbol{\Phi}_{\tau}),
\cos(\boldsymbol{\Phi}_{\tau}),\mathcal{N}_{A}(\mathbf{A}_{\tau}),
\mathbf{U}
\bigr).
\end{aligned}
\label{eq:method-phasenet-input}
\end{equation}
The two branches then predict the raw future amplitude and phase spectra as
\begin{equation}
\begin{aligned}
\bar{\mathbf{A}}_{t+\ell}
&=
\mathcal{N}_{A}^{-1}\left(
\mathcal{N}_{A}(\mathbf{A}_{t})
+
f_{A}(\mathbf{H}_{A})\right),\\
\bar{\boldsymbol{\Phi}}_{t+\ell}
&=
\boldsymbol{\Phi}_{t}
+
f_{\Phi}(\mathbf{H}_{\Phi}).
\end{aligned}
\label{eq:method-branch-prediction}
\end{equation}
Here, \(f_{A}(\cdot)\) and \(f_{\Phi}(\cdot)\) denote AmpliNet and PhaseNet, respectively.

\subsubsection{Image-Prior Spectrum Correction}

\begin{figure}[t]
\centering
\includegraphics[width=\columnwidth]{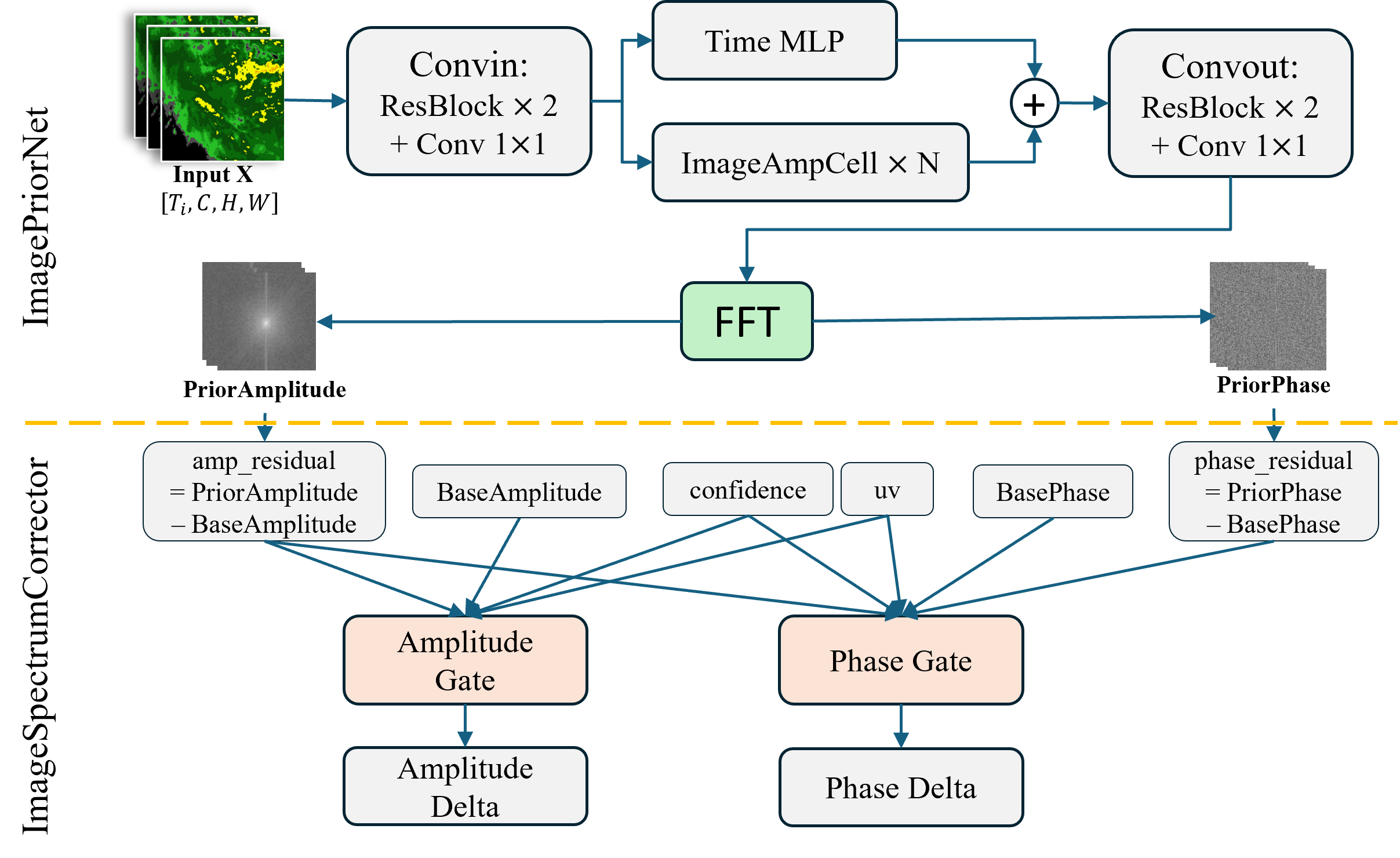}
\caption{Image-prior spectrum correction. ImagePriorNet first generates a future echo prior in the image domain. Based on this prior, ImageSpectrumCorrector conservatively corrects the predicted amplitude and phase spectra.}
\label{fig:imagepriornet_imagespectrumcorrector}
\end{figure}

Frequency-domain prediction alone may miss local morphology cues that are more explicit in the image domain. As illustrated in Fig.~\ref{fig:imagepriornet_imagespectrumcorrector}, we introduce ImagePriorNet and ImageSpectrumCorrector to provide a conservative image-guided correction to the raw spectral prediction. To distinguish the amplitude domains involved in the subsequent operations, we denote the normalized raw amplitude prediction by \(\bar{\mathbf{A}}^{\mathrm n}_{t+\ell} = \mathcal{N}_{A}(\bar{\mathbf{A}}_{t+\ell})\). The superscript \(\mathrm n\) indicates
that an amplitude quantity is represented in the normalized
domain. 

ImagePriorNet first produces an image-domain prior \(\mathbf{P}_{t+\ell}\) from the historical radar echo sequence. The prior is then transformed into the frequency domain as \(\mathbf{Z}^{p}_{t+\ell}=\mathrm{FFT}(\mathbf{P}_{t+\ell})\), with \(\mathbf{A}^{p}_{t+\ell}=|\mathbf{Z}^{p}_{t+\ell}|\) and \(\boldsymbol{\Phi}^{p}_{t+\ell}=\arg(\mathbf{Z}^{p}_{t+\ell})\). The superscript \(\mathrm p\) denotes a quantity derived from
the image-domain prior. The spectral discrepancy between the image prior and the raw branch prediction is computed as
\begin{equation}
\begin{aligned}
\mathbf{R}^{p}_{A,t+\ell}
&=
\mathcal{N}_{A}(\mathbf{A}^{p}_{t+\ell})
-
\bar{\mathbf{A}}^{\mathrm n}_{t+\ell},\\
\mathbf{R}^{p}_{\Phi,t+\ell}
&=
\boldsymbol{\Phi}^{p}_{t+\ell}
-
\bar{\boldsymbol{\Phi}}_{t+\ell}.
\end{aligned}
\label{eq:method-prior-residual}
\end{equation}
Instead of directly replacing the predicted spectra, ImageSpectrumCorrector injects these residuals through learnable gates:
\begin{equation}
\begin{aligned}
\hat{\mathbf{A}}^{0,\mathrm n}_{t+\ell}
&=
\bar{\mathbf{A}}^{\mathrm n}_{t+\ell}
+
\mathbf{G}_{A,t+\ell}
\odot
\mathbf{R}^{p}_{A,t+\ell},\\
\hat{\boldsymbol{\Phi}}^{0}_{t+\ell}
&=
\bar{\boldsymbol{\Phi}}_{t+\ell}
+
\mathbf{G}_{\Phi,t+\ell}
\odot
% \mathbf{C}^{p}_{t+\ell}
\boldsymbol{\Gamma}^{p}_{\Phi,t+\ell}
\odot
\mathbf{R}^{p}_{\Phi,t+\ell},\\
\hat{\mathbf{A}}^{0}_{t+\ell}
&=
\mathcal{N}_{A}^{-1}
(\hat{\mathbf{A}}^{0,\mathrm n}_{t+\ell}).
\end{aligned}
\label{eq:method-gated-correction}
\end{equation}
Here, \(\mathbf{G}_{A,t+\ell}\) and \(\mathbf{G}_{\Phi,t+\ell}\) are amplitude and phase gates, and \(\boldsymbol{\Gamma}^{p}_{\Phi,t+\ell}\) denotes the phase-confidence
map estimated from the image prior. The superscript \(0\) identifies the intermediate spectral estimate produced by the first-stage correction path. This design lets the image prior supplement local morphology while preserving the spectral structure learned by the backbone.

\subsubsection{Spectrum-to-Image Reconstruction and Spatial Refinement}

\begin{figure}[t]
\centering
\includegraphics[width=\columnwidth]{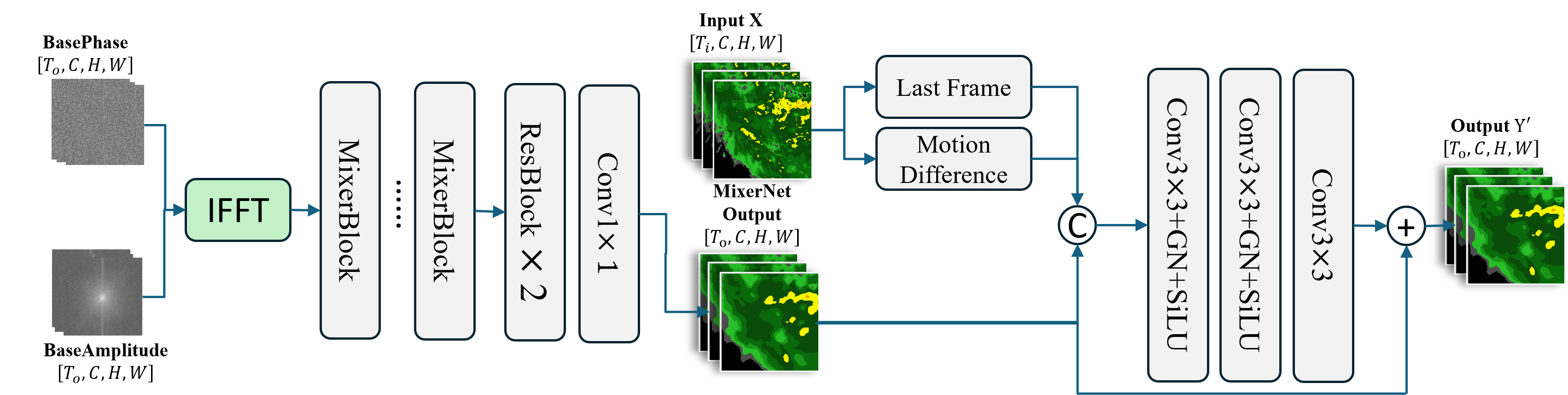}
\caption{Spectrum-to-image reconstruction and spatial refinement. MixerNet further fuses frequency-domain, temporal and spatial information to obtain a more stable future echo estimate. SpatialRefiner is used to perform lightweight residual refinement.}
\label{fig:mixernet_spatialrefiner}
\end{figure}

After spectral correction, the deterministic baseline is reconstructed from the corrected amplitude and phase spectra. The reconstruction and refinement modules are shown in Fig.~\ref{fig:mixernet_spatialrefiner}. The baseline complex spectrum is written as
\begin{equation}
\hat{\mathbf{Z}}^{0}_{t+\ell}
=
\hat{\mathbf{A}}^{0}_{t+\ell}
\odot
\exp
\left(
i\hat{\boldsymbol{\Phi}}^{0}_{t+\ell}
\right).
\label{eq:method-baseline-spectrum}
\end{equation}
The corresponding image-domain prediction is obtained by inverse Fourier transform and further processed by MixerNet:
\begin{equation}
\mathbf{Y}^{m}_{t+\ell}
=
M_{\xi}
\left(
\mathrm{IFFT}
(\hat{\mathbf{Z}}^{0}_{t+\ell})
\right),
\label{eq:method-mixernet}
\end{equation}
where \(M_{\xi}(\cdot)\) denotes MixerNet. MixerNet converts the corrected spectral representation into temporally coherent radar frames by combining frequency, temporal, and spatial mixing operations.

SpatialRefiner then applies a lightweight residual correction in the image domain. It uses the MixerNet prediction, the last observed frame, and the most recent frame difference:
\begin{equation}
\hat{\mathbf{Y}}^{0}
=
\mathbf{Y}^{m}
+
\lambda_{s}
S_{\eta}
\left(
\operatorname{Concat}
[
\mathbf{Y}^{m},
X_{t},
X_{t}-X_{t-1}
]
\right),
\label{eq:method-spatial-refiner}
\end{equation}
where \(S_{\eta}(\cdot)\) denotes SpatialRefiner and \(\lambda_{s}\) controls the residual scale. The output \(\hat{\mathbf{Y}}^{0}\) is used as the deterministic forecast, and \(\bar{\mathbf{A}}^{\mathrm n}_{t+\ell}\) and \(\bar{\boldsymbol{\Phi}}_{t+\ell}\) serve as the spectral baseline for residual diffusion.

The deterministic backbone is trained with image-domain and frequency-domain supervision:
\begin{equation}
\mathcal{L}_{\mathrm{det}}
=
\mathcal{L}_{\mathrm{img}}
+
\lambda_{A}\mathcal{L}_{A}
+
\lambda_{\Phi}\mathcal{L}_{\Phi}
+
\lambda_{P}\mathcal{L}_{P}.
\label{eq:method-det-loss}
\end{equation}
Here, \(\mathcal{L}_{\mathrm{img}}\) denotes pixel-wise MSE, \(\mathcal{L}_{A}\) denotes normalized amplitude-spectrum MSE, \(\mathcal{L}_{\Phi}\) denotes an amplitude-weighted circular phase loss, and \(\mathcal{L}_{P}\) denotes stochastic FACL~\cite{yan2024fourier} regularization of the image-prior branch.

\subsection{Stochastic Amplitude Residual Diffusion}

The deterministic backbone provides a structured prediction, but it may still underestimate or overestimate echo intensity. FreCast therefore models the remaining error as an amplitude residual in the normalized frequency domain. Here, we distinguish the image-prior-corrected intermediate spectra from the spectral baseline used in the diffusion stage. The corrected intermediate spectra \(\hat{\mathbf{A}}^{0,\mathrm{n}}_{t+\ell}\) and \(\hat{\boldsymbol{\Phi}}^{0}_{t+\ell}\) are used to reconstruct the intermediate deterministic forecast \(\hat{\mathbf{Y}}^{0}\). By contrast, the amplitude residual target and diffusion condition are defined using the raw spectral predictions \(\bar{\mathbf{A}}^{\mathrm{n}}_{t+\ell}\) and \(\bar{\boldsymbol{\Phi}}_{t+\ell}\).

\subsubsection{Amplitude Residual Target and Conditioning}

Given the ground-truth future sequence \(Y\), its normalized amplitude spectrum is computed as
\begin{equation}
\mathbf{A}^{\mathrm{gt},\mathrm n}_{t+\ell}
=
\mathcal{N}_{A}
\left(
\left|
\mathrm{FFT}(Y_{t+\ell})
\right|
\right).
\label{eq:method-gt-amplitude}
\end{equation}
The normalized amplitude residual is defined by
\begin{equation}
\Delta\mathbf{A}^{\mathrm n}_{t+\ell}
=
\mathbf{A}^{\mathrm{gt},\mathrm n}_{t+\ell}
-
\bar{\mathbf{A}}^{\mathrm n}_{t+\ell}.
\label{eq:method-normalized-residual}
\end{equation}
It represents the spectral-energy bias left by the deterministic baseline. AmpResidualDM is conditioned on the baseline amplitude and the low-frequency phase structure:
\begin{equation}
\begin{aligned}
\mathbf{C}_{t+\ell}
=
\operatorname{Concat}
\bigl[
&\operatorname{Clip}
(\bar{\mathbf{A}}^{\mathrm n}_{t+\ell}),\cos(\mathcal{T}_{k}(\bar{\boldsymbol{\Phi}}_{t+\ell})),\sin(\mathcal{T}_{k}(\bar{\boldsymbol{\Phi}}_{t+\ell}))
\bigr].
\end{aligned}
\label{eq:method-residual-condition}
\end{equation}
where \(\mathcal{T}_{k}(\cdot)\) denotes low-frequency truncation. The amplitude condition provides the current energy baseline, and the phase condition preserves the dominant spatial arrangement.

\subsubsection{Diffusion-Based Residual Generation}

\begin{figure}[t]
\centering
\includegraphics[width=\columnwidth]{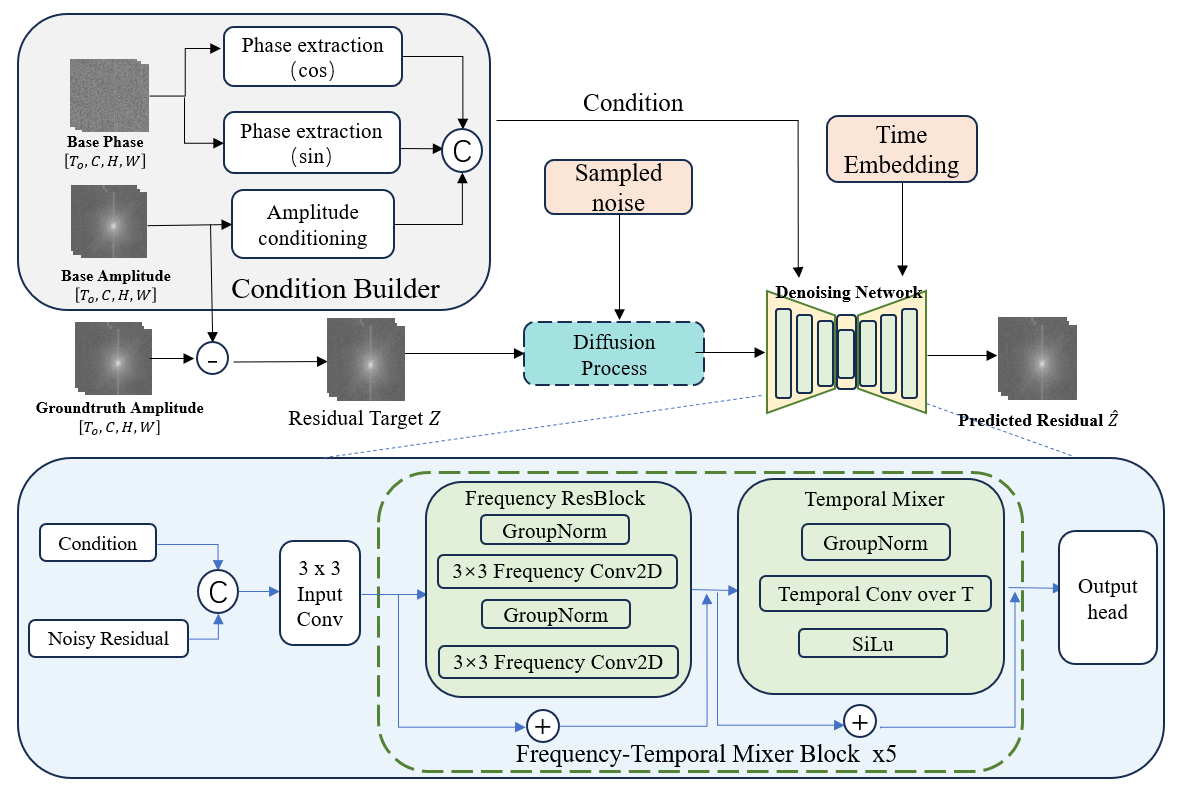}
\caption{Conditional diffusion model for amplitude residual generation.}
\label{fig:amp_residual_dm}
\end{figure}

The architecture of AmpResidualDM is shown in Fig.~\ref{fig:amp_residual_dm}. Let \(\mathbf{z}_{0}=\Delta\mathbf{A}^{\mathrm n}_{t+\ell}\). During training, the forward diffusion process adds Gaussian noise to the residual:
\begin{equation}
\mathbf{z}_{s}
=
\sqrt{\bar{\alpha}_{s}}\mathbf{z}_{0}
+
\sqrt{1-\bar{\alpha}_{s}}\boldsymbol{\epsilon},
\quad
\boldsymbol{\epsilon}
\sim
\mathcal{N}(\mathbf{0},\mathbf{I}),
\label{eq:method-forward-diffusion}
\end{equation}
where \(s\) denotes the diffusion step and \(\bar{\alpha}_{s}\) is determined by the noise schedule. The denoising network \(D_{\omega}\) predicts the injected noise from the noisy residual and the condition:
\begin{equation}
\hat{\boldsymbol{\epsilon}}
=
D_{\omega}
(\mathbf{z}_{s},\mathbf{C}_{t+\ell},s).
\label{eq:method-denoising}
\end{equation}
The diffusion objective is written as
\begin{equation}
\mathcal{L}_{\mathrm{diff}}
=
\mathbb{E}_{s,\boldsymbol{\epsilon}}
\left[
\rho
\left(
\mathbf{W}
\odot
(
\hat{\boldsymbol{\epsilon}}
-
\boldsymbol{\epsilon}
)
\right)
\right],
\label{eq:method-diff-loss}
\end{equation}
where \(\mathbf{W}\) is a frequency-dependent weight map and \(\rho(\cdot)\) is a smooth robust penalty.

\subsubsection{Phase-Preserving Residual Correction}

During inference, AmpResidualDM samples an amplitude residual under the condition \(\mathbf{C}_{t+\ell}\):
\begin{equation}
\Delta\hat{\mathbf{A}}^{\mathrm n}_{t+\ell}
\sim
p_{\omega}
\left(
\Delta\mathbf{A}^{\mathrm n}_{t+\ell}
\mid
\mathbf{C}_{t+\ell}
\right).
\label{eq:method-residual-sampling}
\end{equation}
The corrected amplitude spectrum is obtained by adding the sampled residual to the deterministic baseline:
\begin{equation}
\begin{aligned}
\tilde{\mathbf{A}}^{\mathrm n}_{t+\ell}
&=
\bar{\mathbf{A}}^{\mathrm n}_{t+\ell}
+
\Delta\hat{\mathbf{A}}^{\mathrm n}_{t+\ell},\\
\tilde{\mathbf{A}}_{t+\ell}
&=
\mathcal{N}_{A}^{-1}
(\tilde{\mathbf{A}}^{\mathrm n}_{t+\ell}).
\end{aligned}
\label{eq:method-corrected-amplitude}
\end{equation}
The corrected amplitude is then recombined with the baseline phase:
\begin{equation}
\tilde{\mathbf{Z}}_{t+\ell}
=
\tilde{\mathbf{A}}_{t+\ell}
\odot
\exp
\left(
i\bar{\boldsymbol{\Phi}}_{t+\ell}
\right).
\label{eq:method-final-spectrum}
\end{equation}
The corrected spectrum \(\tilde{\mathbf{Z}}_{t+\ell}\) is subsequently transformed into the image domain and spatially refined following Eqs.~\eqref{eq:method-mixernet} and~\eqref{eq:method-spatial-refiner}, respectively. In Eq.~\eqref{eq:method-mixernet}, \(\tilde{\mathbf{Z}}_{t+\ell}\) replaces \(\hat{\mathbf{Z}}^{0}_{t+\ell}\), and the resulting output is taken as the final forecast \(\tilde{\mathbf{Y}}\). By reusing this reconstruction and refinement path, the stochastic module corrects the amplitude residual while preserving the phase-derived spatial structure estimated by the deterministic backbone.

\begin{figure}[t]
\centering
\includegraphics[width=\columnwidth]{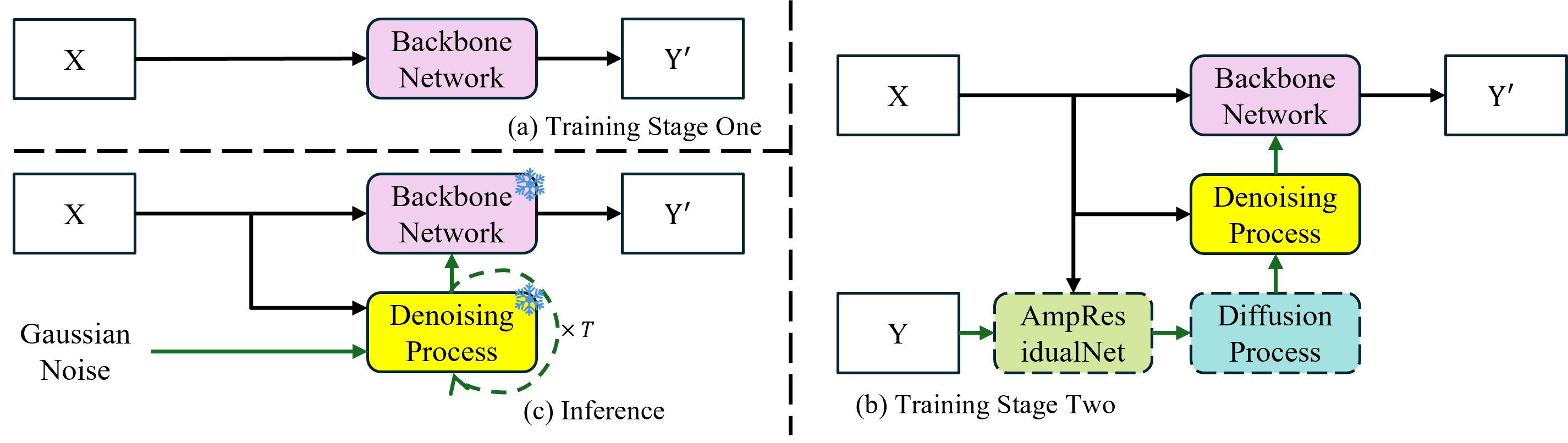}
\caption{Two-stage training and inference strategy of FreCast. FreCast adopts a two-stage training procedure that first learns a deterministic baseline and then models amplitude residual.}
\label{fig:schematic_of_training_inference_strategy}
\end{figure}

\subsection{Training and Inference}

FreCast is trained in two stages, as illustrated in Fig.~\ref{fig:schematic_of_training_inference_strategy}. In the first stage, the deterministic backbone is optimized with \(\mathcal{L}_{\mathrm{det}}\) to produce a stable spectral baseline and an intermediate forecast. In the second stage, the trained backbone is used to construct \(\bar{\mathbf{A}}^{\mathrm n}_{t+\ell}\), \(\bar{\boldsymbol{\Phi}}_{t+\ell}\), and the residual target \(\Delta\mathbf{A}^{\mathrm n}_{t+\ell}\). AmpResidualDM is then trained with \(\mathcal{L}_{\mathrm{diff}}\).

During inference, FreCast first obtains the deterministic baseline from the input sequence. The baseline amplitude and phase are used to build the diffusion condition. AmpResidualDM samples an amplitude residual, which is added to the baseline amplitude spectrum. The corrected amplitude and the retained phase are then recombined and transformed back to the image domain to obtain the final prediction.

\section{Experiments and analyses}
\label{sec:experiments_and_analyses}
\subsection{Experimental Setup}
\textbf{Datasets.} We conduct experiments on three radar precipitation nowcasting datasets: SEVIR~\cite{sevir}, MeteoNet~\cite{meteonet}, and Shanghai Radar~\cite{shanghai}. SEVIR contains radar event sequences over the continental United States from 2017 to 2020, MeteoNet provides radar observations over northwestern France from 2016 to 2018, and Shanghai Radar contains radar echo maps collected in Pudong, Shanghai from October 2015 to July 2018. All radar frames are resized to \(128\times128\), and the input and output sequence lengths are set to \(T_{\mathrm{in}}=5\) and \(T_{\mathrm{out}}=20\), respectively. The dataset splits and input-output settings are summarized in Table~\ref{tab:dataset_settings}. For threshold-based evaluation, we use the standard thresholds of each dataset: [16, 74, 133, 160, 181, 219] for SEVIR, [12, 18, 24, 32] for MeteoNet, and [20, 30, 35, 40] for Shanghai Radar.

\begin{table}[t]
\centering
\caption{Dataset splits and input-output settings.}
\label{tab:dataset_settings}
\resizebox{\columnwidth}{!}{
\begin{tabular}{lcccccc}
\toprule
Dataset & Train & Val & Test & \((C,H,W)\) & \(T_{\mathrm{in}}\) & \(T_{\mathrm{out}}\) \\
\midrule
SEVIR    & 23812 & 6040 & 8106 & \((1,128,128)\) & 5 & 20 \\
MeteoNet & 3102  & 351  & 916  & \((1,128,128)\) & 5 & 20 \\
Shanghai & 2779  & 528  & 528  & \((1,128,128)\) & 5 & 20 \\
\bottomrule
\end{tabular}
}
\end{table}

\textbf{Metrics.} We evaluate image quality using MAE, PSNR, SSIM, and LPIPS, and operational forecasting skill using CSI, HSS, POD, FAR, and FSS. MAE measures absolute intensity error, PSNR measures signal fidelity, SSIM assesses structural similarity, and LPIPS quantifies perceptual discrepancy. CSI jointly accounts for hits, misses, and false alarms, whereas HSS measures forecast skill relative to random forecasts. POD denotes the fraction of observed precipitation events that are detected, while FAR denotes the fraction of predicted events that are false alarms. To account for spatial tolerance and alleviate sensitivity to small displacements, we also report the neighborhood-based FSS. Lower MAE, LPIPS, and FAR values indicate better performance, whereas higher values of the other metrics are preferred. Threshold-dependent metrics are computed at dataset-specific precipitation-intensity thresholds and then averaged across thresholds.
    
\textbf{Baselines.} We compare FreCast with six representative nowcasting baselines under the same data splits and evaluation protocol. PhyDNet~\cite{phydnet}, SimVP~\cite{simvp}, EarthFormer~\cite{earthformer}, and AlphaPre~\cite{alphapre} are deterministic methods, while DiffCast~\cite{diffcast} and NowcastNet~\cite{nowcastnet} are probabilistic methods. In Table~\ref{tab:image_quality_metrics}, PhyDNet, SimVP, and EarthFormer are categorized as deterministic non-decoupled methods. AlphaPre is categorized as a deterministic decoupled method because it explicitly predicts amplitude and phase components. DiffCast and NowcastNet are categorized as probabilistic decoupled methods because they decompose precipitation evolution into deterministic and stochastic components. FreCast is also categorized as a probabilistic decoupled method, but differs by preserving the phase-related spatial structure and modeling only the amplitude residual for intensity correction.

\textbf{Training details.} FreCast is trained in two stages. The amplitude, phase, and image-prior terms are weighted by \(\lambda_A=0.01\), \(\lambda_\Phi=0.02\), and \(\lambda_P=0.02\), respectively. In the second stage, the backbone is frozen and only the conditional amplitude-residual diffusion model is trained. Both stages use the AdamW optimizer with a learning rate of \(1\times10^{-4}\) and \(\beta_1=0.9\) and \(\beta_2=0.95\). FreCast and all baseline models were trained for 100 epochs. All experiments are conducted on a workstation equipped with an Intel(R) Xeon(R) Silver 4310 CPU @ 2.10 GHz, two NVIDIA RTX 5880 Ada Generation GPUs with 48 GB memory each, and 256 GB RAM.

\begin{table}[t]
\centering
\caption{Image quality metrics on three radar nowcasting datasets. Bold values indicate the best performance for each metric. Det. denotes deterministic models, Prob. denotes probabilistic generative models, Non-dec. denotes non-decoupled models, and Dec. denotes decoupled models. The yellow and green backgrounds distinguish deterministic and probabilistic generative models, respectively.}
\label{tab:image_quality_metrics}
\renewcommand{\arraystretch}{1.06}
\setlength{\tabcolsep}{2.1pt}
\resizebox{\columnwidth}{!}{%
\begin{tabular}{c c c c c c c}
\toprule
\textbf{Dataset} & \textbf{Method} & \textbf{Type} &
\textbf{$\downarrow$MAE} & \textbf{$\uparrow$SSIM} &
\textbf{$\downarrow$LPIPS} & \textbf{$\uparrow$PSNR} \\
\midrule
\multirow{7}{*}{SEVIR}
& \detc{PhyDNet} & \detc{Det./Non-dec.} & \detc{8.2103} & \detc{0.6878} & \detc{0.3272} & \detc{27.3824} \\
& \detc{SimVP} & \detc{Det./Non-dec.} & \detc{\textbf{8.1634}} & \detc{0.6879} & \detc{0.3668} & \detc{\textbf{27.6084}} \\
& \detc{EarthFormer} & \detc{Det./Non-dec.} & \detc{8.4129} & \detc{0.6831} & \detc{0.3831} & \detc{26.9942} \\
& \detc{AlphaPre} & \detc{Det./Dec.} & \detc{8.3659} & \detc{\textbf{0.6970}} & \detc{0.2835} & \detc{27.4611} \\
& \probc{DiffCast} & \probc{Prob./Dec.} & \probc{9.6737} & \probc{0.6597} & \probc{\textbf{0.1775}} & \probc{25.7700} \\
& \probc{NowcastNet} & \probc{Prob./Dec.} & \probc{9.7630} & \probc{0.6531} & \probc{0.2434} & \probc{25.8882} \\
& \probc{FreCast} & \probc{Prob./Dec.} & \probc{9.4507} & \probc{0.6472} & \probc{0.2819} & \probc{27.2283} \\
\midrule
\multirow{7}{*}{MeteoNet}
& \detc{PhyDNet} & \detc{Det./Non-dec.} & \detc{2.0264} & \detc{0.6880} & \detc{0.3717} & \detc{27.9125} \\
& \detc{SimVP} & \detc{Det./Non-dec.} & \detc{\textbf{1.8196}} & \detc{\textbf{0.7401}} & \detc{0.2729} & \detc{28.3307} \\
& \detc{EarthFormer} & \detc{Det./Non-dec.} & \detc{2.3655} & \detc{0.5144} & \detc{0.3373} & \detc{27.6379} \\
& \detc{AlphaPre} & \detc{Det./Dec.} & \detc{1.8414} & \detc{0.6970} & \detc{0.2572} & \detc{\textbf{29.1892}} \\
& \probc{DiffCast} & \probc{Prob./Dec.} & \probc{1.8731} & \probc{0.7361} & \probc{\textbf{0.1272}} & \probc{27.7334} \\
& \probc{NowcastNet} & \probc{Prob./Dec.} & \probc{2.3518} & \probc{0.7081} & \probc{0.2188} & \probc{26.2025} \\
& \probc{FreCast} & \probc{Prob./Dec.} & \probc{1.9747} & \probc{0.6469} & \probc{0.2579} & \probc{28.9150} \\
\midrule
\multirow{7}{*}{Shanghai}
& \detc{PhyDNet} & \detc{Det./Non-dec.} & \detc{1.9217} & \detc{0.7289} & \detc{0.3350} & \detc{26.5654} \\
& \detc{SimVP} & \detc{Det./Non-dec.} & \detc{1.7290} & \detc{0.7806} & \detc{0.2699} & \detc{26.5376} \\
& \detc{EarthFormer} & \detc{Det./Non-dec.} & \detc{1.8122} & \detc{0.7536} & \detc{0.3190} & \detc{26.5418} \\
& \detc{AlphaPre} & \detc{Det./Dec.} & \detc{\textbf{1.6700}} & \detc{0.7756} & \detc{0.2728} & \detc{\textbf{27.2903}} \\
& \probc{DiffCast} & \probc{Prob./Dec.} & \probc{1.7383} & \probc{\textbf{0.7881}} & \probc{\textbf{0.1442}} & \probc{25.9726} \\
& \probc{NowcastNet} & \probc{Prob./Dec.} & \probc{2.2019} & \probc{0.7661} & \probc{0.2254} & \probc{24.5629} \\
& \probc{FreCast} & \probc{Prob./Dec.} & \probc{1.6852} & \probc{0.7677} & \probc{0.2126} & \probc{27.1556} \\
\bottomrule
\end{tabular}%
}
\end{table}

\begin{table}[t]
\centering
\caption{Operational nowcasting skill on three radar nowcasting datasets. Bold values indicate the best performance for each metric. Det. denotes deterministic models, Prob. denotes probabilistic generative models, Non-dec. denotes non-decoupled models, and Dec. denotes decoupled models. The yellow and green backgrounds distinguish deterministic and probabilistic generative models, respectively.}
\label{tab:operational_nowcasting_metrics}
\renewcommand{\arraystretch}{1.12}
\setlength{\tabcolsep}{2.0pt}
\resizebox{\columnwidth}{!}{%
\begin{tabular}{c c c c c c c c}
\toprule
\textbf{Dataset} & \textbf{Method} & \textbf{Type} &
\textbf{$\uparrow$CSI} & \textbf{$\uparrow$HSS} &
\textbf{$\uparrow$POD} & \textbf{$\downarrow$FAR} &
\textbf{$\uparrow$FSS} \\
\midrule
\multirow{7}{*}{SEVIR}
& \detc{PhyDNet} & \detc{Det./Non-dec.} & \detc{0.3315} & \detc{0.4154} & \detc{0.3979} & \detc{0.3994} & \detc{0.5998} \\
& \detc{SimVP} & \detc{Det./Non-dec.} & \detc{0.3243} & \detc{0.4084} & \detc{0.3889} & \detc{\textbf{0.3099}} & \detc{0.6018} \\
& \detc{EarthFormer} & \detc{Det./Non-dec.} & \detc{0.3328} & \detc{0.4240} & \detc{0.4039} & \detc{0.3939} & \detc{0.5854} \\
& \detc{AlphaPre} & \detc{Det./Dec.} & \detc{0.3350} & \detc{0.4253} & \detc{0.4016} & \detc{0.4334} & \detc{0.6209} \\
& \probc{DiffCast} & \probc{Prob./Dec.} & \probc{0.3206} & \probc{0.4168} & \probc{0.4264} & \probc{0.5500} & \probc{0.6221} \\
& \probc{NowcastNet} & \probc{Prob./Dec.} & \probc{0.3102} & \probc{0.3999} & \probc{0.4102} & \probc{0.5456} & \probc{0.6167} \\
& \probc{FreCast} & \probc{Prob./Dec.} & \probc{\textbf{0.3489}} & \probc{\textbf{0.4440}} & \probc{\textbf{0.4532}} & \probc{0.4839} & \probc{\textbf{0.6414}} \\
\midrule
\multirow{7}{*}{MeteoNet}
& \detc{PhyDNet} & \detc{Det./Non-dec.} & \detc{0.3812} & \detc{0.4956} & \detc{0.4640} & \detc{0.3474} & \detc{0.5708} \\
& \detc{SimVP} & \detc{Det./Non-dec.} & \detc{0.4106} & \detc{0.5306} & \detc{0.4963} & \detc{0.3488} & \detc{0.6281} \\
& \detc{EarthFormer} & \detc{Det./Non-dec.} & \detc{0.4303} & \detc{0.5568} & \detc{0.5780} & \detc{0.4101} & \detc{0.6490} \\
& \detc{AlphaPre} & \detc{Det./Dec.} & \detc{0.4417} & \detc{0.5624} & \detc{0.5212} & \detc{\textbf{0.2995}} & \detc{0.6442} \\
& \probc{DiffCast} & \probc{Prob./Dec.} & \probc{0.4016} & \probc{0.5220} & \probc{0.5065} & \probc{0.4025} & \probc{0.6475} \\
& \probc{NowcastNet} & \probc{Prob./Dec.} & \probc{0.4072} & \probc{0.5293} & \probc{\textbf{0.6444}} & \probc{0.4981} & \probc{0.6405} \\
& \probc{FreCast} & \probc{Prob./Dec.} & \probc{\textbf{0.4475}} & \probc{\textbf{0.5701}} & \probc{0.5518} & \probc{0.3298} & \probc{\textbf{0.6612}} \\
\midrule
\multirow{7}{*}{Shanghai}
& \detc{PhyDNet} & \detc{Det./Non-dec.} & \detc{0.3964} & \detc{0.5286} & \detc{0.4571} & \detc{0.2708} & \detc{0.3679} \\
& \detc{SimVP} & \detc{Det./Non-dec.} & \detc{0.4427} & \detc{0.4667} & \detc{0.5353} & \detc{0.3227} & \detc{0.5683} \\
& \detc{EarthFormer} & \detc{Det./Non-dec.} & \detc{0.4555} & \detc{0.5986} & \detc{0.5708} & \detc{0.3349} & \detc{0.5453} \\
& \detc{AlphaPre} & \detc{Det./Dec.} & \detc{0.4417} & \detc{0.5777} & \detc{0.5085} & \detc{\textbf{0.2702}} & \detc{0.5346} \\
& \probc{DiffCast} & \probc{Prob./Dec.} & \probc{0.4167} & \probc{0.5539} & \probc{0.5111} & \probc{0.3507} & \probc{0.5448} \\
& \probc{NowcastNet} & \probc{Prob./Dec.} & \probc{0.4214} & \probc{0.5606} & \probc{\textbf{0.6235}} & \probc{0.4340} & \probc{0.5496} \\
& \probc{FreCast} & \probc{Prob./Dec.} & \probc{\textbf{0.4674}} & \probc{\textbf{0.6055}} & \probc{0.5696} & \probc{0.3113} & \probc{\textbf{0.5901}} \\
\bottomrule
\end{tabular}%
}
\end{table}

\subsection{Experimental Results}
To evaluate the effectiveness of FreCast for precipitation nowcasting, we compared FreCast with other baseline algorithms on three datasets. We report both image quality metrics and operational forecasting metrics. This comparison helps determine whether a model only generates visually similar radar echo images or genuinely improves the prediction of precipitation events, strong echo structures, and spatial neighborhood consistency.

Table~\ref{tab:image_quality_metrics} reports the image quality metrics of different methods on the three datasets, including MAE, SSIM, LPIPS, and PSNR. Deterministic models still show clear advantages in pixel-level reconstruction metrics, especially MAE and PSNR. This result is consistent with the optimization behavior of deterministic models, which tend to produce locally smooth predictions with small average pixel deviations under regression losses. However, lower pixel error does not necessarily indicate better precipitation nowcasting. Reasonable echo displacement can be strongly penalized by MAE and PSNR, while smoothed strong echo cores may still obtain favorable average error values. The LPIPS results further support this observation. As a perceptual image quality metric, LPIPS shows that probabilistic generative models outperform deterministic models. DiffCast achieves the best LPIPS on all three datasets, indicating that pixel-domain generative refinement can produce local details with more realistic perceptual textures. However, such visual details do not necessarily correspond to more accurate precipitation event prediction.

Table~\ref{tab:operational_nowcasting_metrics} reports the operational forecasting metrics of different methods on the three datasets, including CSI, HSS, POD, FAR, and FSS. FreCast achieves the best CSI, HSS, and FSS on all three datasets. This indicates that FreCast improves the overall discrimination of precipitation events, the skill score relative to random forecasts, and spatial neighborhood consistency. The POD and FAR results reveal a typical tradeoff in generative nowcasting models. NowcastNet achieves the highest POD on MeteoNet and Shanghai, but its FAR is also substantially higher than that of FreCast. This suggests that NowcastNet tends to expand the predicted precipitation area to increase the hit rate. In contrast, FreCast maintains a high POD while keeping FAR lower than those of DiffCast and NowcastNet. These results indicate that frequency-domain amplitude residual refinement can enhance precipitation detection while imposing a more effective constraint on spurious echo expansion.

\begin{figure}[t]
\centering
\includegraphics[width=\columnwidth]{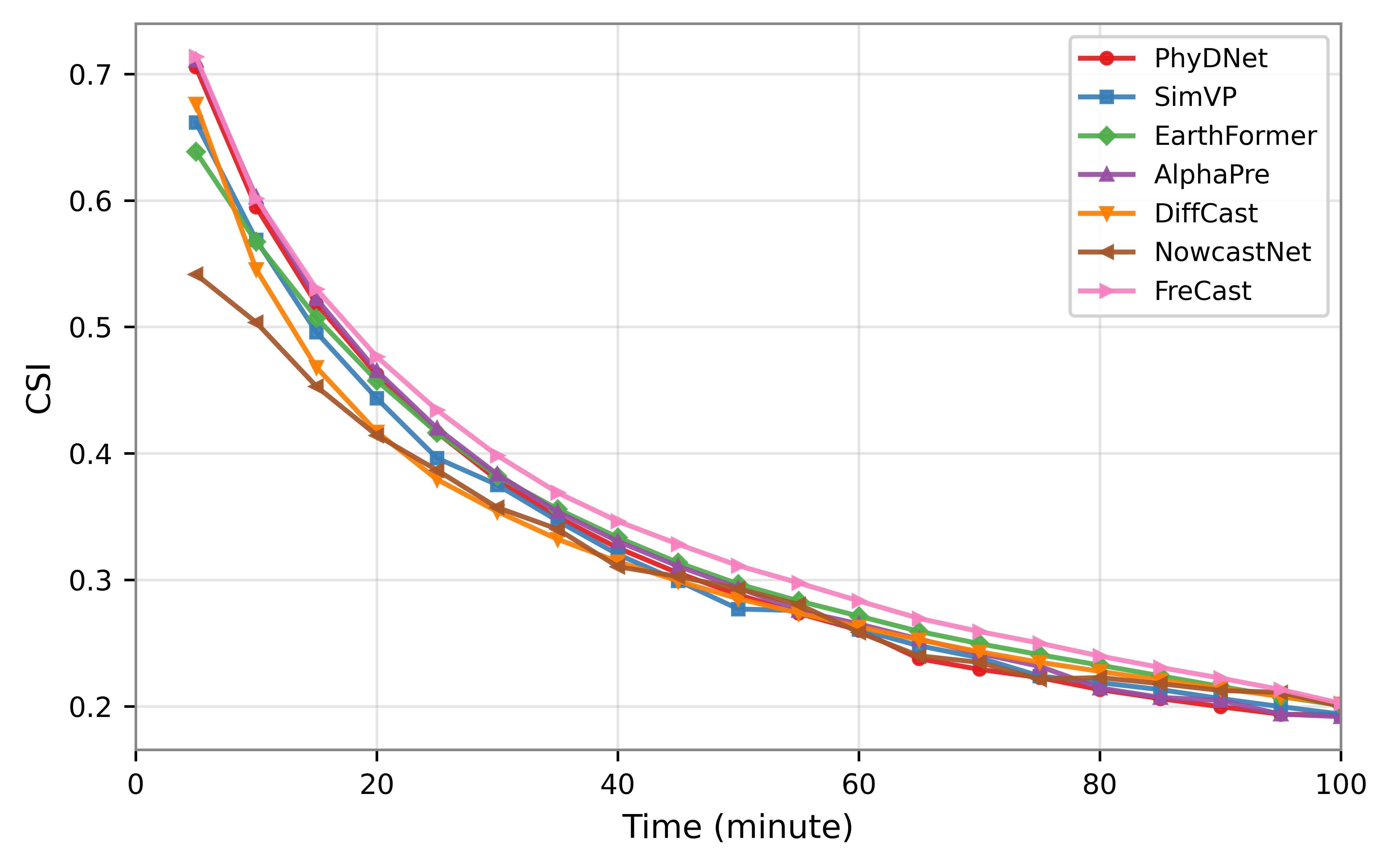}
\caption{Variation of CSI with forecasting lead time.}
\label{fig:p0_methods_CSI_curve_compare}
\end{figure}

\begin{figure}[t]
\centering
\includegraphics[width=\columnwidth]{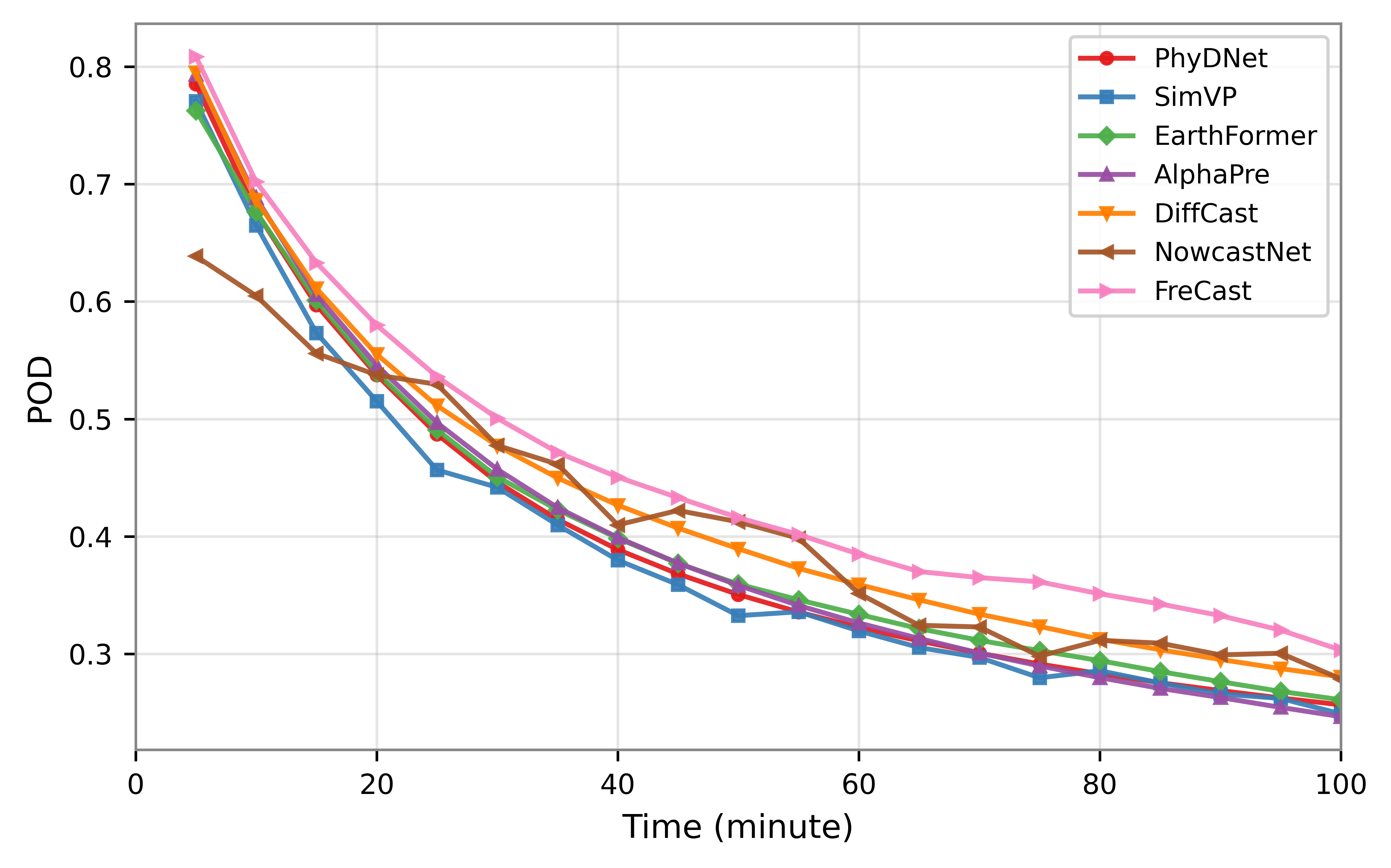}
\caption{Variation of POD with forecasting lead time.}
\label{fig:p0_methods_POD_curve_compare}
\end{figure}

Fig.~\ref{fig:p0_methods_CSI_curve_compare} and Fig.~\ref{fig:p0_methods_POD_curve_compare} further show the variations in CSI and POD over different forecast lead times on the SEVIR dataset. As the forecast lead time increases from 10 min to 100 min, the CSI and POD of all methods gradually decrease. This indicates that the uncertainty in future echo position, morphology, and intensity increases with lead time. Nevertheless, FreCast maintains the highest or nearly highest CSI at most lead times and preserves a higher POD curve at medium and long lead times. This suggests that FreCast can effectively propagate existing echo structures at short lead times and still maintain strong precipitation event discrimination after forecast uncertainty accumulates. DiffCast and NowcastNet, as probabilistic generative models, show weaker CSI curves than FreCast overall. This indicates that simply generating more precipitation structures does not necessarily improve operational forecasting performance. These results are consistent with the design objective of frequency-domain amplitude residual refinement, which enhances future echo representation through generative modeling while applying constrained correction to the energy distribution. This reduces structural drift and unstable predictions as the forecast lead time increases.

\begin{figure}[t]
\centering
\includegraphics[width=\columnwidth]{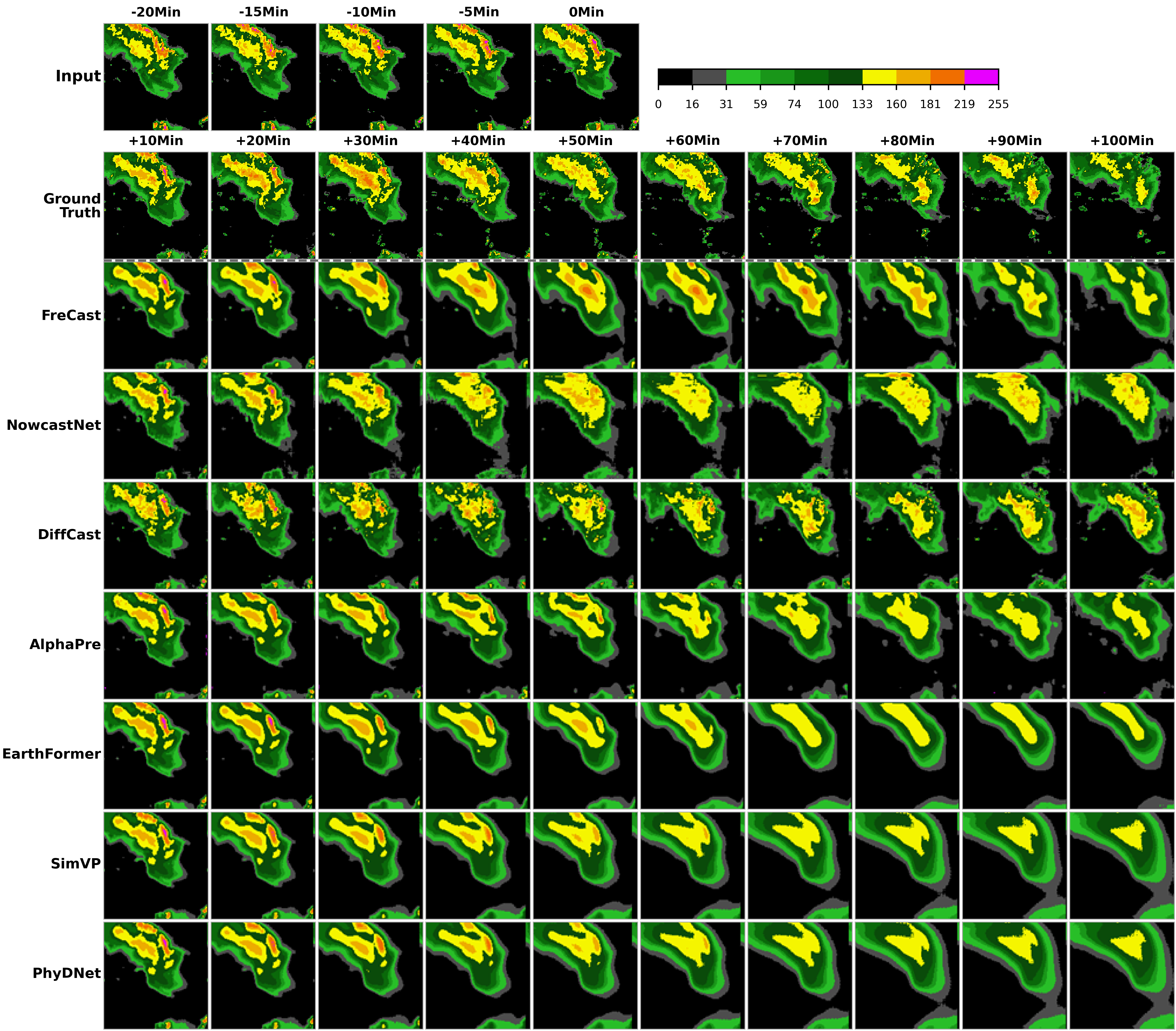}
\caption{Qualitative comparison of precipitation nowcasting results.}
\label{fig:p0_methods_prediction_visualization_compare}
\end{figure}

Fig.~\ref{fig:p0_methods_prediction_visualization_compare} presents a qualitative prediction example for a precipitation event from the SEVIR dataset. Deterministic models usually generate spatially continuous but overly smooth echo structures. As the forecast lead time increases, the high-intensity cores gradually weaken. Pixel-domain probabilistic generative methods can recover sharper local textures, but they are also more prone to fragmented structures, positional instability, and additional spurious echoes. FreCast better recovers the yellow and orange strong echo regions while preserving the geometric pattern and motion trend of the main rainband.

\subsection{Analysis and discussions}

To further analyze the sources of the performance gains achieved by the proposed model, we formulate three questions around the model design and conduct experiments for each question.

\begin{figure}[t]
\centering
\includegraphics[width=\columnwidth]{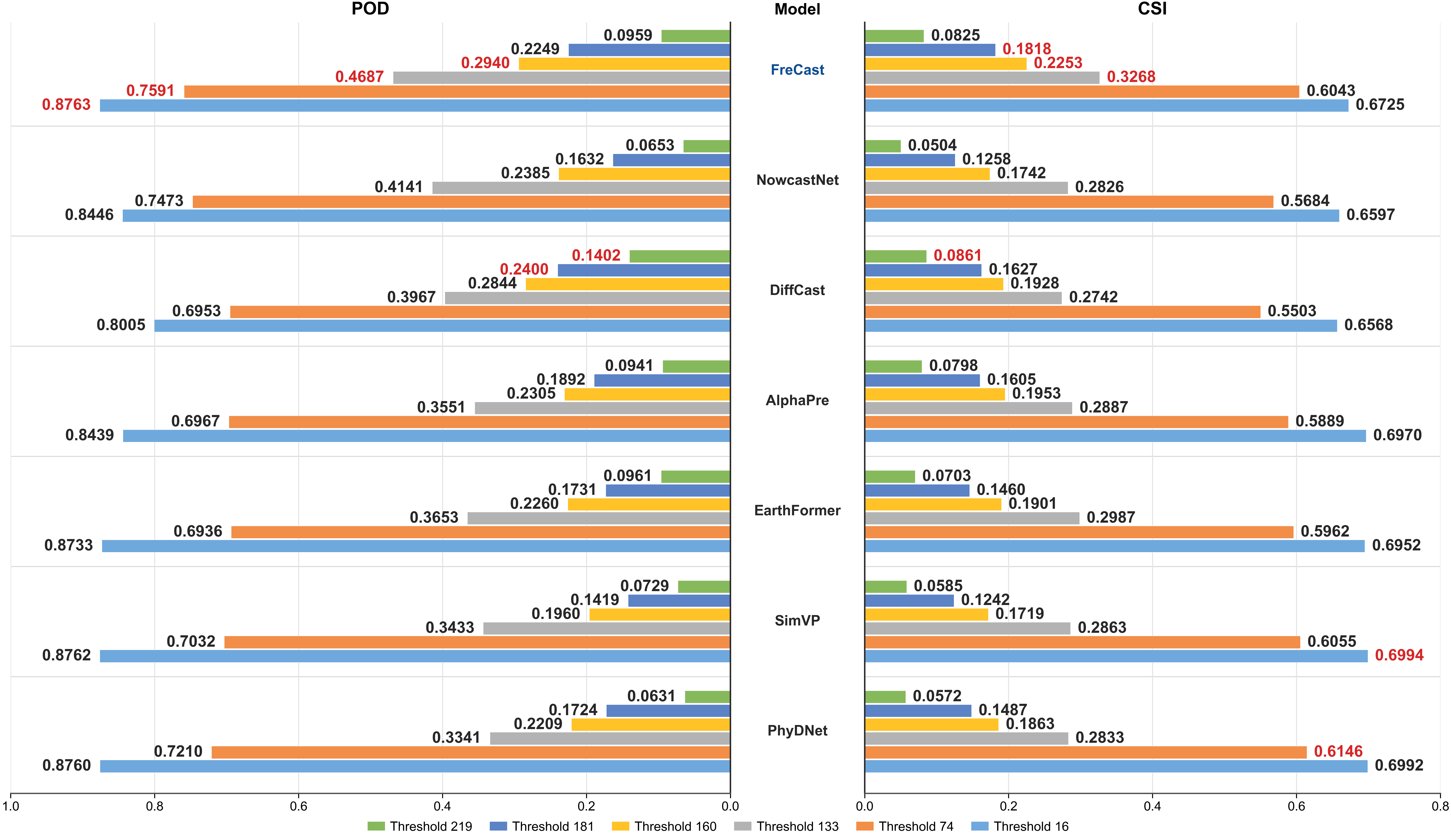}
\caption{Comparison of POD and CSI under six echo intensity thresholds. POD (left) and CSI (right) are arranged around the central model column. the POD axis is reversed so that longer outward-extending bars indicate better performance on both sides. Colors denote thresholds 16, 74, 133, 160, 181, and 219, and red values mark the best score among all methods at each threshold.}
\label{fig:p1_methods_CSI_POD-Threshold_compare}
\end{figure}

\subsubsection{Forecasting Performance Under Different Echo Intensity Thresholds} To verify whether FreCast maintains stable advantages across different radar echo intensity levels and to further examine its ability to correct the underestimation of strong echoes, we compare the CSI and POD of different methods under multiple echo intensity thresholds on the SEVIR dataset. CSI-Threshold evaluates the overall discrimination ability of the model for precipitation events at different intensities, while POD-Threshold measures the hit rate for observed precipitation events at different intensities.

Fig.~\ref{fig:p1_methods_CSI_POD-Threshold_compare} show the CSI and POD results under different radar echo intensity thresholds on the SEVIR dataset. As the threshold increases from 16 to 219, the CSI and POD of all methods decrease markedly. This indicates that strong echo regions are sparser, smaller in spatial scale, and more difficult to predict in terms of growth, decay, and displacement. Therefore, the results at high thresholds better reflect the ability of a model to characterize strong precipitation structures.

In terms of CSI, FreCast shows more pronounced advantages at medium and high intensity thresholds. It achieves CSI values of 0.3268, 0.2253, and 0.1818 at CSI-133, CSI-160, and CSI-181, respectively, all of which are the best among the compared methods. At the extremely high threshold CSI-219, DiffCast is slightly better than FreCast. This result suggests that pixel-domain generative refinement may be more inclined to generate stronger or broader local echoes in extremely strong echo regions, thereby improving hit-related metrics at very high thresholds. However, FreCast consistently leads at medium and high thresholds and remains close to the best result at the extremely high threshold. This indicates that frequency-domain amplitude residual refinement does not merely improve large-scale weak echoes, but can also more specifically alleviate the underestimation of strong echoes. The POD results further support this conclusion. FreCast achieves POD values of 0.8763, 0.7591, 0.4687, and 0.2940 at POD-16, POD-74, POD-133, and POD-160, respectively, all of which are the best results. This shows that FreCast maintains a higher event hit rate across multiple levels ranging from weak echoes to relatively strong echoes. Meanwhile, DiffCast achieves higher POD at POD-181 and POD-219. Combined with the FAR results discussed above, the higher POD at extremely high thresholds may be associated with more aggressive generation of high-intensity echoes, whereas FreCast emphasizes a better balance among hit rate, false alarm control, and spatial consistency.

\begin{figure}[t]
\centering
\includegraphics[width=\columnwidth]{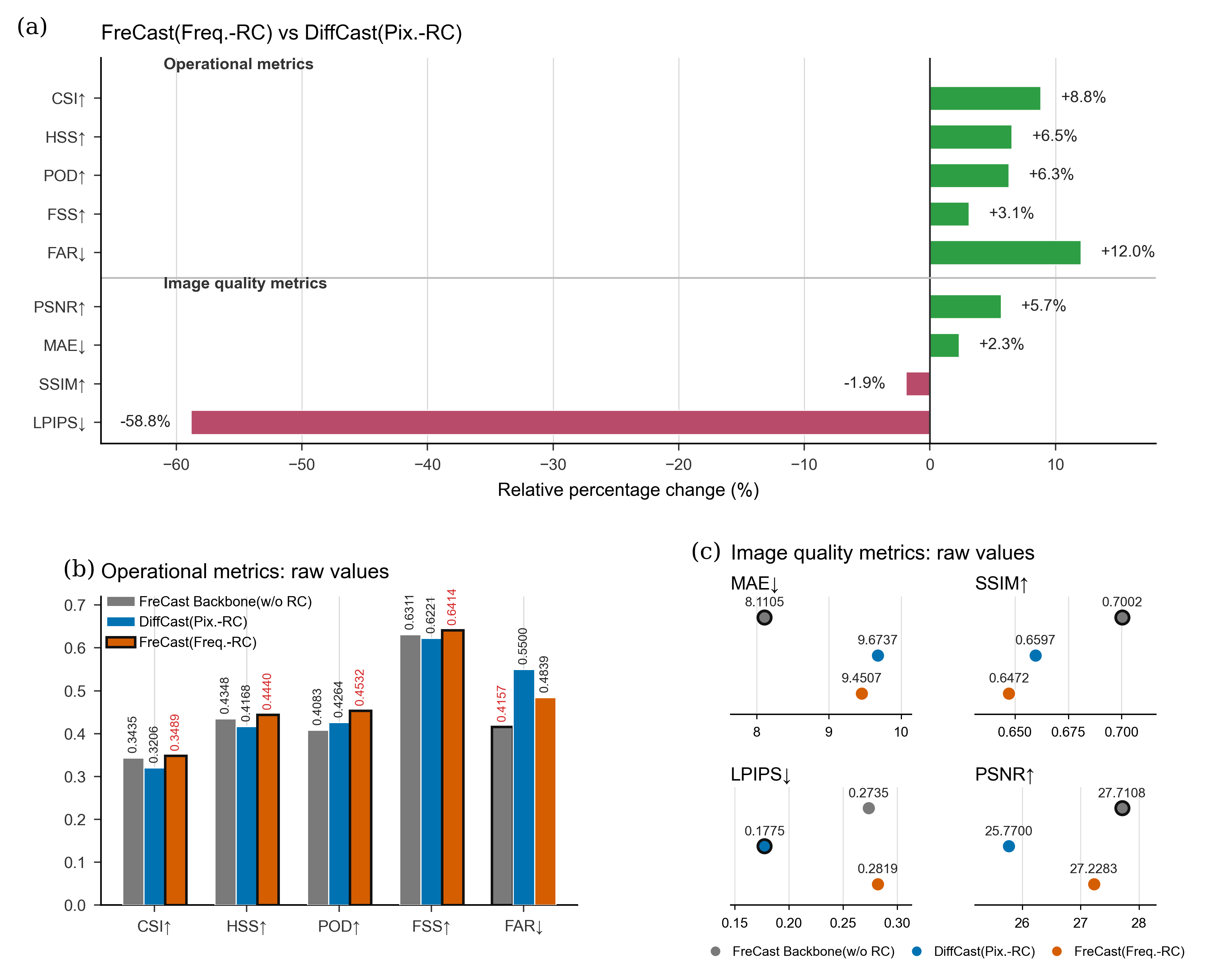}
\caption{Ablation analysis of residual refinement. \textbf{(a)}, Direction-corrected relative changes using DiffCast as the reference, where positive values indicate performance improvement. \textbf{(b)}, Raw operational nowcasting metrics, where red values indicate the best performance for each metric. \textbf{(c)}, Raw image quality metrics.}
\label{fig:p2_frecast_amplitude_residual_repair_compare}
\end{figure}

\begin{figure}[t]
\centering
\includegraphics[width=\columnwidth]{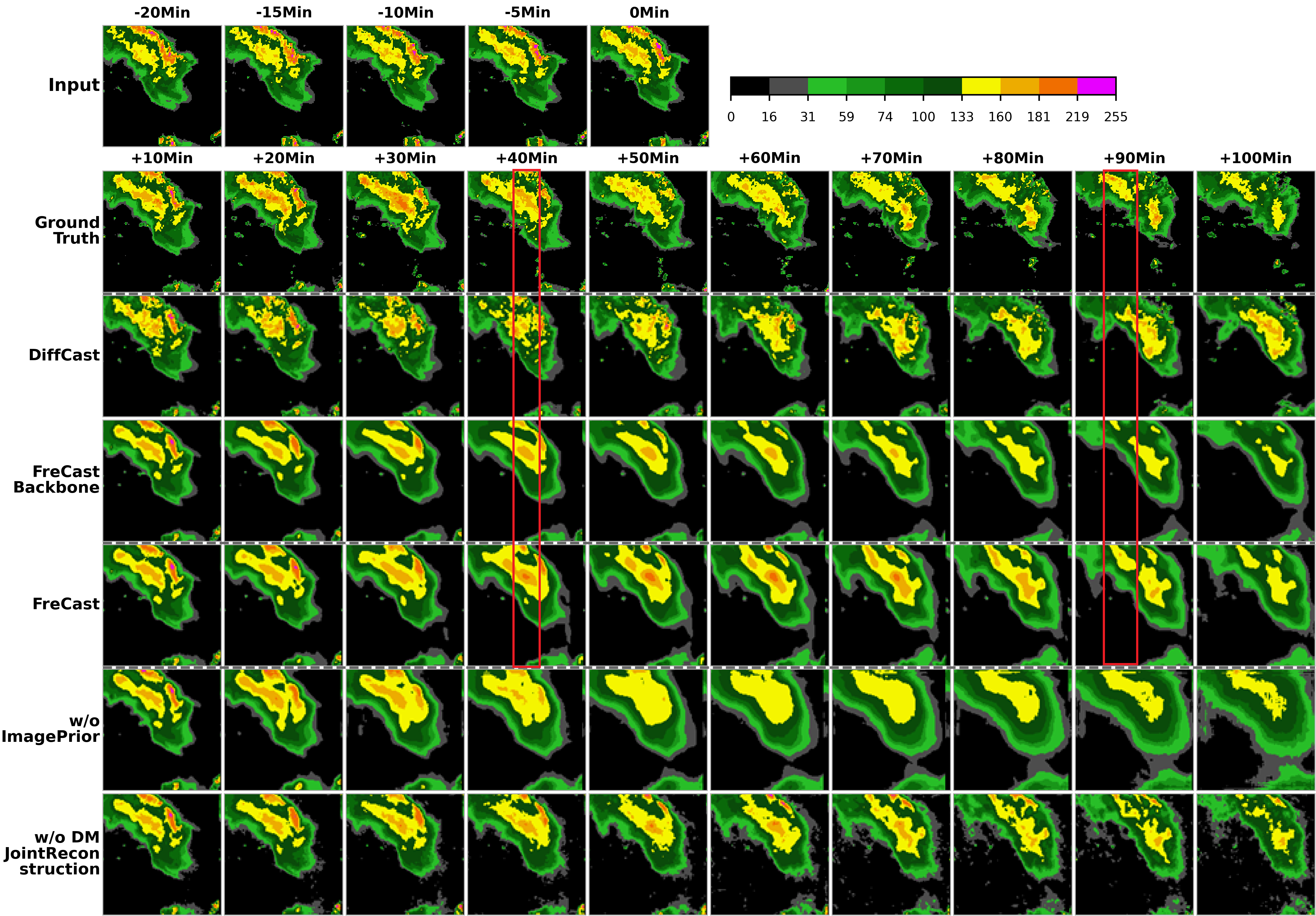}
\caption{Qualitative comparisons of frequency-domain amplitude residual refinement and key FreCast component ablations. The upper rows compare DiffCast, the FreCast Backbone, and FreCast to illustrate differences in the recovery of strong echoes and rainband structure preservation among residual refinement strategies. Red boxes highlight representative regions with strong echo. The lower rows compare FreCast with the w/o ImagePrior and w/o DM Joint Reconstruction variants to assess the effects of these two components on the predictions.}
\label{fig:p2_and_p3_frecast_ablation_prediction_visualization_compare}
\end{figure}

\subsubsection{Effectiveness of Frequency-Domain Amplitude Residual Refinement} To verify the role of residual correction, we compare the FreCast Backbone without residual refinement, the pixel-domain residual refinement version DiffCast, and the frequency-domain amplitude residual refinement version FreCast. This comparison examines whether residual refinement compensates for echo-intensity underestimation and whether amplitude correction better preserves spatial structure than pixel-domain residual generation.

Fig.~\ref{fig:p2_frecast_amplitude_residual_repair_compare}(a) shows that, compared with DiffCast, FreCast achieves improvements in the desired directions across all operational forecasting metrics, including higher CSI, HSS, POD, and FSS, and lower FAR. This indicates that frequency-domain amplitude residual refinement not only improves the hit rate of precipitation events, but also improves false alarm control. Fig.~\ref{fig:p2_frecast_amplitude_residual_repair_compare}(b) reports the raw operational metrics of the three variants. The complete FreCast achieves the best CSI, HSS, POD, and FSS among the three methods. The FAR of the Backbone is 0.4157, which is lower than the 0.4839 of FreCast. This indicates that the deterministic baseline without residual refinement is more conservative and produces fewer spurious echoes, but this conservativeness is associated with lower POD and weaker recovery of strong echoes. Compared with the FAR of 0.5500 obtained by DiffCast, FreCast substantially reduces the false alarm ratio. This suggests that phase-preserving amplitude refinement can reduce the unconstrained structural expansion introduced by residual generation in the pixel-domain. Fig.~\ref{fig:p2_frecast_amplitude_residual_repair_compare}(c) shows the tradeoff in image quality metrics. The Backbone performs best in MAE, SSIM, and PSNR, while DiffCast achieves the best LPIPS. FreCast does not pursue the best performance across all image quality metrics. Compared with DiffCast, FreCast improves PSNR and MAE but obtains a worse LPIPS, indicating that frequency-domain refinement sacrifices part of the perceptual texture score. More importantly, this tradeoff leads to more stable CSI, HSS, POD, and FSS performance. 

The qualitative ablation results in the upper part of Fig.~\ref{fig:p2_and_p3_frecast_ablation_prediction_visualization_compare}, comparing DiffCast, the FreCast Backbone, and FreCast, further explain these observations. The red boxes highlight the predictions of strong-echo regions produced by the different models. The FreCast Backbone can preserve the approximate morphology of the main precipitation region, but its predictions are overly smooth and the strong echo cores are clearly weakened. DiffCast generates more local textures and high-intensity patches through pixel-domain residual refinement, but the local structures are more fragmented and accompanied by spurious echoes. In contrast, the complete FreCast better recovers the yellow and orange strong echoes within the red-box region while preserving the spatial position and overall morphology of the main rainband.

\begin{figure}[t]
\centering
\includegraphics[width=\columnwidth]{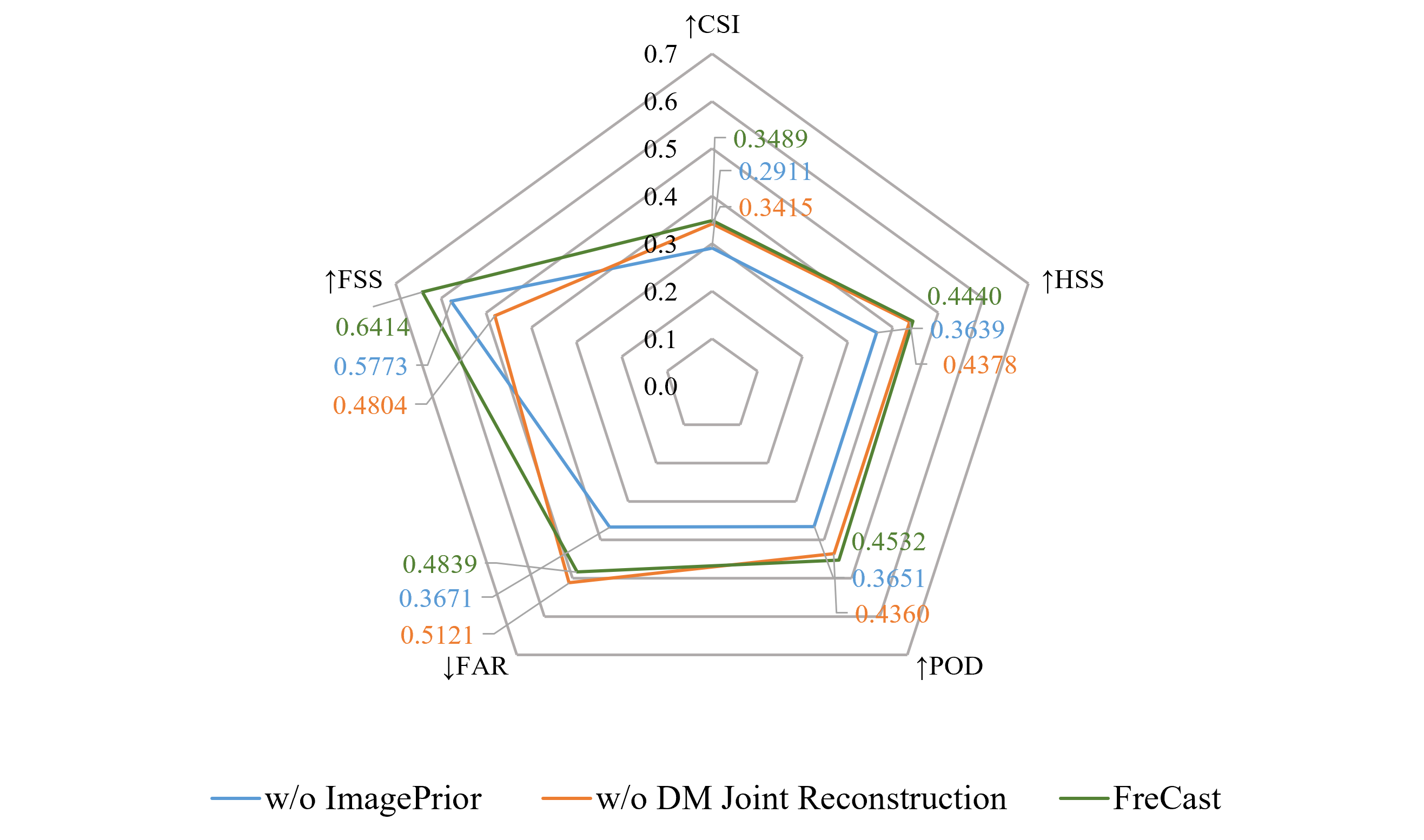}
\caption{Ablation analysis of key FreCast components.}
\label{fig:p3_frecast_component_ablation_compare}
\end{figure}

\subsubsection{Ablation Analysis of Model Components and the Joint Reconstruction Strategy} To further analyze the contribution of each component in the complete FreCast, we constructed two ablation variants, w/o ImagePrior and w/o DM Joint Reconstruction, and compare them with the complete FreCast. The w/o ImagePrior variant removes both ImagePriorNet and ImageSpectrumCorrector. The w/o DM Joint Reconstruction variant combines the sampled amplitude residual with the amplitude spectrum and phase spectrum of the frozen backbone network, but does not use the diffusion-backbone joint reconstruction path. Instead, it directly reconstructs the prediction from the corrected spectrum.

Fig.~\ref{fig:p3_frecast_component_ablation_compare} compares the performance of the complete FreCast with the two variants. The complete FreCast achieves the best CSI, HSS, POD, and FSS, indicating that ImagePrior and DM Joint Reconstruction make complementary contributions to event recognition, intensity recovery, and spatial consistency. After removing ImagePrior, the CSI, HSS, POD, and FSS decrease to 0.2911, 0.3639, 0.3651, and 0.5773, respectively. Although the FAR of this variant decreases to 0.3671, this mainly reflects a more conservative prediction pattern rather than stronger overall forecasting ability. The lower FAR is accompanied by lower CSI and POD, indicating that the model reduces spurious echoes but also weakens its ability to capture true precipitation events. This suggests that ImagePrior provides necessary spatial structural priors for frequency-domain prediction and helps the model recover more plausible local echo morphology and strong echo regions. When DM Joint Reconstruction is not used, the CSI, HSS, and POD of the model are close to those of the complete FreCast, but FSS decreases to 0.4804 and FAR increases to 0.5121. This indicates that direct reconstruction from the corrected spectrum can preserve part of the event hit capability, but the absence of the joint reconstruction constraint weakens spatial neighborhood consistency and increases unreliable local echoes. 

The qualitative ablation results in the lower part of Fig.~\ref{fig:p2_and_p3_frecast_ablation_prediction_visualization_compare}, comparing FreCast, w/o ImagePrior, and w/o DM Joint Reconstruction, are consistent with the above interpretation. The w/o ImagePrior variant predicts a smoother main precipitation band and insufficiently recovers the strong echo cores and boundary details. The w/o DM Joint Reconstruction variant preserves more local textures, but it tends to produce scattered echoes and unstable boundaries. The complete FreCast achieves a better balance among strong echo recovery, rainband continuity, and structural stability.

Overall, the performance improvement of FreCast comes from three interrelated designs. First, the model uses a deterministic spectral backbone to provide a phase-related spatial structural anchor. Second, amplitude residual diffusion restricts generative refinement to the correction of intensity and scale-wise energy distribution. Third, ImagePrior and DM Joint Reconstruction jointly constrain the corrected spectrum, enabling the model to recover strong echoes while preserving rainband morphology and temporal stability.

\section{Conclusion}
\label{sec:conclusion}

This study re-examines multi-step precipitation nowcasting through prediction-error analysis. When models proposed in prior studies can already capture the locations and morphologies of the main precipitation regions reasonably well, echo-intensity bias remains an important source of error. We therefore formulate nowcasting as the evolution and correction of radar-echo intensity fields and propose FreCast.

FreCast decomposes radar echoes into amplitude and phase, retains the backbone phase as a structural anchor, and models frequency-domain amplitude residuals with conditional diffusion. Across the evaluated radar datasets, FreCast reduced echo-intensity errors and improved nowcasting performance, consistent with stronger echo recovery and less structural disruption.

 However, Its effectiveness depends on the quality of the backbone phase; amplitude-only refinement cannot fully correct substantial displacement or morphological errors. Future work will investigate joint amplitude–phase correction and incorporate multisource observations and physical constraints to improve reliability and interpretability in complex precipitation scenarios.

% \section*{Acknowledgments}
% This should be a simple paragraph before the References to thank those individuals and institutions who have supported your work on this article.

% {\appendix[Proof of the Zonklar Equations]
% Use $\backslash${\tt{appendix}} if you have a single appendix:
% Do not use $\backslash${\tt{section}} anymore after $\backslash${\tt{appendix}}, only $\backslash${\tt{section*}}.
% If you have multiple appendixes use $\backslash${\tt{appendices}} then use $\backslash${\tt{section}} to start each appendix.
% You must declare a $\backslash${\tt{section}} before using any $\backslash${\tt{subsection}} or using $\backslash${\tt{label}} ($\backslash${\tt{appendices}} by itself
%  starts a section numbered zero.)}

% %{\appendices
% %\section*{Proof of the First Zonklar Equation}
% %Appendix one text goes here.
% % You can choose not to have a title for an appendix if you want by leaving the argument blank
% %\section*{Proof of the Second Zonklar Equation}
% %Appendix two text goes here.}

\bibliographystyle{IEEEtran}
\nocite{*}  
\bibliography{references}

\end{document}